\documentclass[lettersize,journal]{IEEEtran}
\usepackage{graphicx}
\usepackage{url}
\usepackage{colortbl}
\usepackage{xcolor}
\usepackage{float}
\usepackage{cuted}
\usepackage[section]{placeins}
\usepackage{soul}
\usepackage{amsmath}
\usepackage{amssymb}
\usepackage{booktabs}
\usepackage{multirow}
\usepackage{tabularx}
\usepackage{array}
\usepackage{makecell}
\usepackage{capt-of}
\providecommand{\keywords}[1]{%
        \par\addvspace\baselineskip%
        \noindent\textbf{\textit{Keywords---}} #1%
}

\title{Energy Estimation for Virtual Servers Using Resource Utilization and Machine Learning}

\author{Amandip Sangha
\thanks{E-mail: amandip.s.sangha@gmail.com}
\thanks{Revised preprint --- 22 October 2026.
This work has been submitted to the IEEE for possible
publication. Copyright may be transferred without
notice, after which this version may no longer be accessible.}}

\date{}

\begin{document}

\maketitle

\begin{abstract}
Estimating the energy use of virtual servers is difficult because
physical power interfaces and privileged host counters are typically accessible
only on the underlying physical host. We investigate a machine learning
approach that predicts the CPU-package power attributable to an individual
virtual server from system and performance metrics observable within that
virtual server itself. During training, host-side RAPL measurements provide the power
target used to learn this mapping, while deployed inference requires only
measurements available within the virtual server. We evaluate the approach
across heterogeneous CPU platforms, multiple workload families, unseen
workloads and hardware, concurrent workload combinations, and varying
resource allocations using temporally blocked evaluation. Prediction is
generally strong when deployment conditions are represented during training,
while zero-shot transfer to unseen workloads, hardware, and configurations is
less reliable. A pooled multi-platform model retains good performance when
target platforms are represented during training, and pooled
multi-configuration modeling performs well across different processor and
memory allocations. Feature-importance analysis reveals interpretable
resource signatures consistent with workload behavior across CPU, memory,
storage, and network activity. Monitoring and inference add little direct CPU-power overhead. The results support practical virtual server power estimation when training data cover the
intended deployment domain.
\end{abstract}

 
\section{Introduction}

\subsection{Motivation and Problem Setting}

\begin{figure}[!ht]
 \includegraphics[width=\columnwidth]{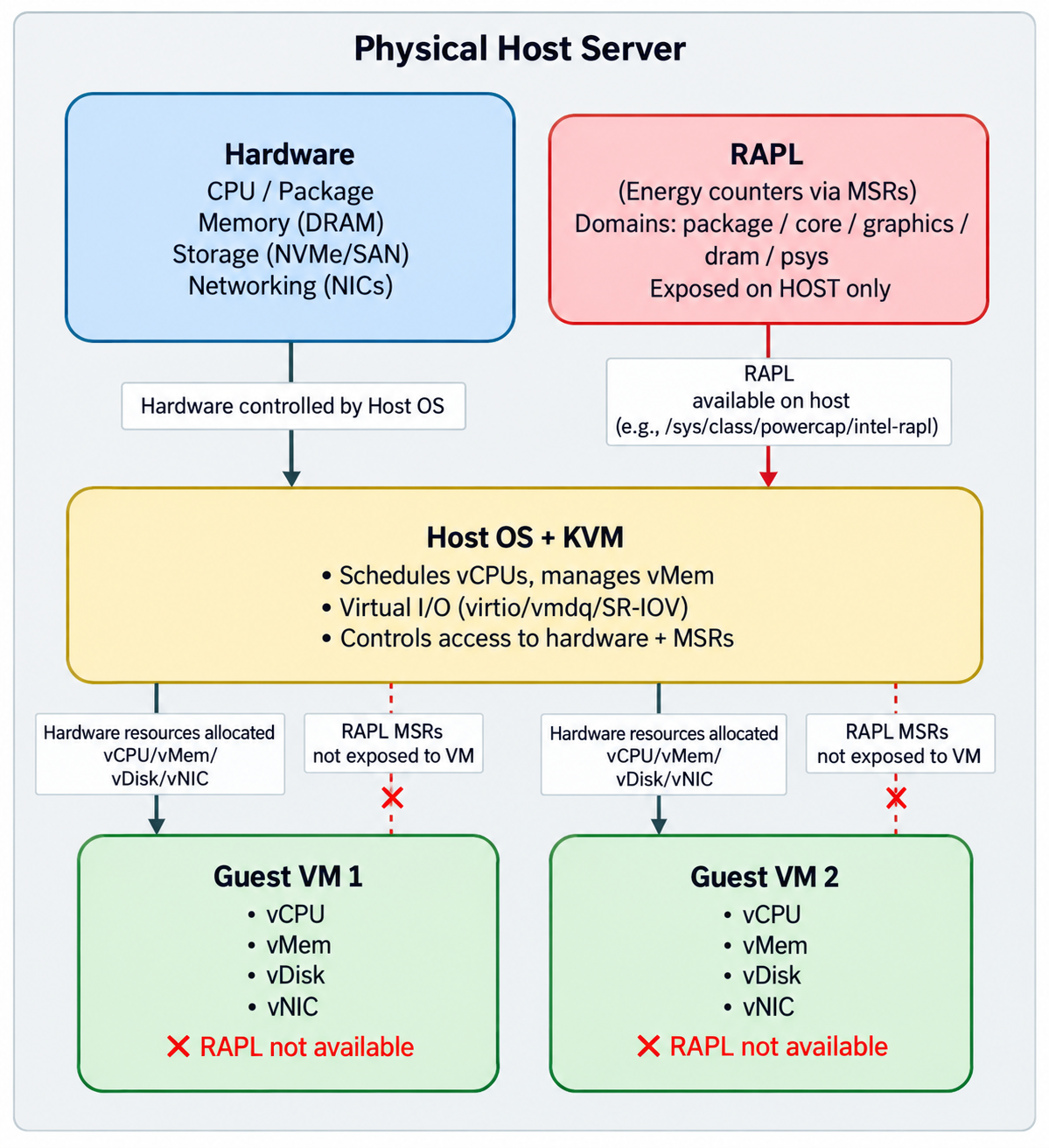}
 \caption{Illustration of host--guest resource allocation in KVM virtualization and how host power interfaces such as RAPL are inaccessible to ordinary guests.}
 \label{fig:hostguestmodel}
\end{figure}

Virtualization is ubiquitous in the IT-sector for server workload isolation, maximization of hardware utilization and improvement of flexibility in configuration, portability, scaling and deployment. The data center and cloud computing industries are examples of fields that would be difficult without virtualization, and thus the infrastructural foundation of most other fields is powered, at least in part, by virtualization. There are different approaches to virtualization, e.g. hypervisor based, kernel based, containers, and so on. The main objective, however, is to provide through some management layer on the physical host, e.g. a hypervisor, common hardware resources such as CPU, RAM, storage, networking and so on, to a guest virtual server, also commonly called a virtual machine (VM). This resource allocation comes with deliberate limits and security constraints: not all hardware interfaces and mechanisms are passed on to or exposed from the physical host to the virtual guest. Figure~\ref{fig:hostguestmodel} illustrates this host--guest
separation in KVM, including the allocation of virtualized resources and the
fact that host RAPL interfaces are not directly exposed to ordinary guest VMs.

In this work, we study how the power consumption associated with a virtual server can be estimated without access to the underlying physical power measurement interfaces of the host. Accurate power and energy estimation at the VM level is relevant for energy-aware scheduling, carbon-footprint reporting, cost optimization, and feedback to cloud tenants. However, direct hardware power interfaces and privileged host counters are typically unavailable from inside a guest VM. This motivates estimation based on guest-side resource measurements, including time-varying CPU utilization, memory usage, storage and network I/O, and related performance metrics observable from within the guest operating system. The practical objective is to enable a VM operator or tenant to estimate the power
and energy associated with their VM during operation, without requiring
privileged access to physical host power interfaces or cooperation from the
hypervisor. This is particularly relevant in cloud environments, where tenants
typically control their virtual machines but not the underlying hardware. Such
an estimator can provide a continuously updated, workload-responsive estimate
derived from the VM's actual resource activity, rather than relying on a
coarse attribution of host-level power according to allocated resources or
fixed proportional shares.

Global data center electricity consumption is increasing, having reached approximately 415 TWh in 2024, equivalent to around 1.5\% of global electricity use. It is projected to reach approximately 950 TWh by 2030 \cite{iea2025energyai, iea2026keyquestions}. This growth increases the importance of accurate energy accounting, cost management, and efficient operation of computing infrastructure.

Improved methods for estimating energy consumption can contribute to this by
making energy consumption more observable at the level where computational workloads
are actually executed. More precise workload-level estimates can support energy-aware operation and
provide a clearer basis for comparing the energy demands of computational
workloads. In virtualized environments, this requires attribution at the level
of individual VMs rather than relying only on aggregate host-level measurements
or coarse allocation rules. The present work addresses this problem by learning
VM-level power estimates from measurements available within the guest operating
system, with host-side power measurements required only during model
construction.

\subsection{Fundamental Research Question}
The present study is organized around the following fundamental question: \textit{Under what conditions can resource utilization measurements observable from within a virtual server be used to estimate its power consumption accurately, and how well does such an estimation model generalize across workloads, hardware platforms, resource configurations, and concurrent workload compositions?}

The question therefore concerns not only prediction accuracy within a given deployment setting, but also the extent to which the learned relationship remains valid across changes in workload, hardware, resource allocation, and workload composition.

\subsection{Research Questions}

The study addresses six research questions (RQ):

\begin{description}
    \item[\textbf{RQ1}] \textbf{Within-platform power estimation:} How accurately can guest-observable resource utilization measurements estimate VM-associated power when training and evaluation are performed on the same hardware platform and workload environment?

    \item[\textbf{RQ2}] \textbf{Workload generalization:} How does estimation performance depend on workload type, and can models generalize to previously unseen workloads?

    \item[\textbf{RQ3}] \textbf{Hardware generalization:} How well do guest-side models transfer across processor models and architectures, including to previously unseen hardware?

    \item[\textbf{RQ4}] \textbf{Concurrent workload generalization:} Can models learned from individual workloads estimate power accurately when workloads execute concurrently?

    \item[\textbf{RQ5}] \textbf{VM configuration sensitivity:} How robust is the learned guest-metric-to-power relationship to changes in vCPU and memory allocation?

    \item[\textbf{RQ6}] \textbf{Runtime overhead and deployability:}
    What resource and energy overhead is introduced by guest-side
    metric collection and ML model inference?
\end{description}

\subsection{Contributions}

The main contributions of this work are:

\begin{itemize}

\item We formulate and evaluate a machine-learning approach for estimating
virtual-server CPU-package power from resource-utilization and performance
metrics observable within the virtual server. Model construction uses
synchronized host-side RAPL measurements as the training target, while
deployed inference requires only measurements available within the virtual
server.

\item We conduct a heterogeneous experimental study across Intel and AMD
platforms and multiple workload families, evaluating within-domain prediction
as well as generalization to previously unseen workloads and hardware.

\item We show that a single pooled model can retain strong performance across
multiple represented hardware platforms, and that pooled multi-configuration
training similarly supports accurate estimation across different vCPU and RAM
allocations.

\item We extend the evaluation to concurrent workloads and analyze
feature importance across platforms and workloads, showing interpretable
resource signatures and quantifying how predictive importance is distributed
across CPU, memory, storage, and network activity.

\item We use temporally blocked evaluation with exclusion gaps to reduce temporal train--test leakage, repeated holdouts to assess stability across test intervals, leave-one-workload-out and leave-one-hardware-out experiments to evaluate workload and hardware generalization, and paired comparisons where appropriate for controlled comparisons.

\end{itemize}

\section{Related Work}
Prior work on energy estimation in virtualized environments has used
external power meters, processor energy interfaces such as RAPL, and
software- or performance-counter measurements to characterize host and
virtual-machine power consumption. VM-level approaches have also used
statistical and machine-learning models, with differing requirements for
host-side instrumentation, hypervisor-visible measurements, privileged
performance counters, and calibration data.

\subsection{Host-Level Server Power and Energy Modeling}

Server power can be characterized either by direct measurement or by estimation
from hardware and software activity. External power meters provide direct
whole-system measurements at the physical server level, but do not by themselves resolve
how that consumption should be attributed to individual processes or virtual
machines. Early server power-modeling work therefore explored relationships
between measured AC power and operating-system or performance-counter activity;
for example, full-system power has been predicted from low-overhead utilization metrics \cite{economou2006fullsystem}.

Modern processors expose on-chip energy counters. Intel introduced the
Running Average Power Limit (RAPL) interface to provide energy measurements for
processor power domains \cite{david2010rapl,rotem2012power}. Subsequent studies
have examined its use for fine-grained energy measurement and validated its
relationship to external power measurements, while also identifying practical
limitations related to supported domains, update behavior, and platform
dependence \cite{hahnel2012measuring,khan2018rapl}.

Measurement, attribution, and prediction should therefore be distinguished.
RAPL provides processor-domain energy measurements, whereas software tools can
attribute measured energy to processes or other software entities using
additional activity information. PowerJoular and Joular Core, for example, combine platform
power measurements with software-level monitoring for process-level power
attribution \cite{noureddine2022powerjoular, noureddine2026joularcore}. Prediction goes one step further by
learning a mapping from observable activity metrics to power, allowing power to
be estimated when the original hardware power interface is unavailable. 

\subsection{VM-Level Power Estimation}

Estimating power at VM granularity is more difficult than measuring host-level
power because a virtual machine has no independent physical power interface.
Early work such as Joulemeter addressed this by learning resource-specific
power models from host instrumentation and combining them with
hypervisor-visible VM resource usage to infer per-VM energy consumption
\cite{kansal2009joulemeter}. This approach therefore depends on host-side
instrumentation and hypervisor support during operation, rather than on
measurements available solely inside the guest.

Other work has used processor performance-monitoring counters to construct
CPU- and memory-power models and then allocate the resulting energy estimates
to individual VMs. Per-VM energy accounting in shared virtualized systems has
also been demonstrated using performance monitoring counter (PMC)-based models
under dynamic voltage and frequency scaling (DVFS)
\cite{bertran2012energy}. Such methods can achieve fine-grained attribution,
but require access to low-level counters and privileged virtualization-layer
information.

The above approaches establish that VM-level power can be inferred from resource
activity, but they typically assume visibility from the host or hypervisor.
The present work considers the complementary deployment setting in which
host-side measurements are available only during model construction, whereas
subsequent guest-side power estimation, i.e. model inference, can be performed entirely from within the guest using only guest-observable metrics.

\subsection{Guest-Side and Resource-Based Estimation}

Several studies have estimated power from software-visible resource activity
rather than from dedicated power meters. VM-level power has been modeled from
resource-usage variables such as CPU, cache, memory, and disk activity, although
the required measurements may be collected from the physical host and intended
for provider-side monitoring \cite{bohra2010vmeter}. Server-level power has
similarly been modeled from CPU, memory, disk, and network utilization using
external power measurements during calibration, with the resulting models tied
to particular hardware configurations \cite{smith2012cloudmonitor}.

Operating-system resource variables, including processor, memory, disk, and
network activity, have also been shown to support machine-learning-based power
estimation across different hardware configurations
\cite{povoa2017resource}. This line of work demonstrates that software-visible
resource activity can provide useful predictive information without requiring
direct power measurements at inference time. More broadly, reviews of power
modeling for virtualized entities emphasize sensitivity to workload,
hardware architecture, virtualization, concurrency, and attribution assumptions,
which limits the plausibility of a single universally transferable power model
\cite{depasquale2023dynamics}.

\subsection{Machine Learning for Power Modeling}

Machine-learning approaches to power estimation have progressed from linear
regression toward nonlinear regression, ensemble methods, and temporal models.
At the VM level, nonlinear support-vector regression using
performance-monitoring counters has been shown to outperform linear models
under heterogeneous workload behavior \cite{veni2016prediction}. At the
server level, neural-network models have been used to capture temporal
dependencies in power traces and improve prediction across heterogeneous
hardware and workload conditions \cite{wu2021elman}.

More recent work has explicitly considered distribution shifts across server
platforms. Heterogeneous server power prediction has been formulated as a
domain-adaptation regression problem, demonstrating that transfer across
hardware can remain challenging even when prediction performance is strong
within the hardware domains represented during training
\cite{mo2025crossserver}. These findings highlight the importance of evaluating
power-estimation models not only within their training domains, but also under
changes in workload and hardware.
 
\subsection{Comparison with Existing Approaches}

The present study differs from prior work primarily in its deployment model,
rather than in the use of a new regression algorithm. Earlier VM-level
approaches infer or attribute power from resource activity, but commonly rely
on host-side instrumentation, hypervisor-visible measurements, or privileged
performance counters during operation
\cite{kansal2009joulemeter,bohra2010vmeter,bertran2012energy}. In contrast,
the present approach requires host-side RAPL-derived power measurements only
during model construction; after training, inference uses resource-utilization
and performance metrics observable from within the virtual server itself. Provider-side power reporting can offer centralized monitoring and potentially more direct measurements, but requires infrastructure integration and provider cooperation to expose VM-level estimates. Our approach instead enables independent guest-side estimation after model construction, without privileged host access during inference.

The study also differs from server-level resource-based approaches. Previous
work has shown that CPU, memory, disk, and network activity can support
power prediction and that calibration can reduce dependence on direct power
measurement at inference time
\cite{smith2012cloudmonitor,povoa2017resource}. However, these approaches
principally estimate physical-server power, whereas the present study targets
VM-attributed CPU-package power and evaluates inference from measurements
available within an ordinary guest VM.

A further distinction concerns evaluation across deployment domains.
Machine-learning studies have demonstrated nonlinear power prediction,
temporal modeling, and cross-server adaptation
\cite{veni2016prediction,wu2021elman,mo2025crossserver}, but the present
evaluation considers workload, hardware, concurrent workload composition,
and VM resource allocation within a common experimental framework. In
particular, leave-one-workload-out and leave-one-hardware-out experiments
measure zero-shot transfer, while pooled multi-platform and
multi-configuration experiments test whether a common model remains effective
when the relevant deployment domains are represented during training.

In the present work, we distinguish and evaluate within-domain prediction from transfer across workloads, hardware platforms, concurrent or mixed workload compositions, and VM resource configurations, and account also for the temporal structure of the measurements. We measure monitoring overhead directly across all six hardware platforms to quantify the practical cost of collecting the guest observable metrics required by the estimator.

Tables~S1--S2 in the supplemental material compare representative approaches in terms of prediction targets, deployment requirements, input features, and reported accuracy. The accuracy figures are not directly comparable because the studies differ in measurement targets, hardware, workloads, and evaluation protocols. 
\section{System Model and Measurement Methodology}
\subsection{Host-Side Power Measurement and VM Attribution}

For model construction, host-side CPU energy is measured through the Linux
RAPL powercap interface using Joular Core \cite{noureddine2026joularcore}. Because RAPL reports energy for
shared physical processor domains rather than for individual VMs, VM-level
attribution is performed on the host before the resulting power measurements
are used as the regression target.

The experiments use KVM virtualization on Linux, with libvirt and virsh for VM
management. Joular Core \cite{noureddine2026joularcore} provides the
host-side monitoring interface. Measurements are recorded at approximately
1~s intervals and associated with the host-side QEMU/KVM process corresponding to the
guest VM.

The RAPL hierarchy was inspected through /sys/class/powercap/. On the Intel
hosts, package-0 was exposed together with core and DRAM subdomains. The AMD
hosts exposed package-0 and core, but no separate DRAM subdomain. Joular
Core's Linux CPU backend uses the RAPL package energy counter as its power
source and computes package power from consecutive cumulative energy readings.
The same package-level RAPL source was therefore used across Intel and AMD
platforms; separately exposed subdomains were not combined into the
measurement target.

For VM-level attribution, the PID of the host-side QEMU/KVM process
corresponding to the guest VM was supplied to Joular Core in process-monitoring
mode. Joular Core obtains system-wide CPU utilization from /proc/stat and
process utilization from /proc/[PID]/stat. Before attribution, an idle
CPU-package baseline is calibrated automatically from five one-second samples.
For package power \(P_{\mathrm{pkg}}(t)\), idle baseline
\(P_{\mathrm{idle}}\), process CPU utilization \(u_{\mathrm{proc}}(t)\), and
system CPU utilization \(u_{\mathrm{sys}}(t)\), the attributed VM power is

\[
P_{\mathrm{VM}}(t)
=
\max\{0,P_{\mathrm{pkg}}(t)-P_{\mathrm{idle}}\}
\frac{u_{\mathrm{proc}}(t)}{u_{\mathrm{sys}}(t)}.
\]

The resulting quantity is sampled at approximately 1~s intervals and corresponds to
``Process Power (W)'' in the Joular Core output. Throughout this paper,
\(P_{\mathrm{VM}}(t)\) denotes VM-attributed CPU-package power: the
idle-baseline-corrected portion of RAPL package power attributed to the
host-side QEMU/KVM process representing the guest VM. It is not a measurement
of whole-system wall power and does not include separately measured power from
components such as storage devices, network interfaces, or DRAM.

\subsection{Host--Guest Learning Model}
Let $\mathbf{x}_t \in \mathbb{R}^{d}$ denote the vector of metrics observable from inside the guest VM at
time $t$. We define the regression target as
\begin{equation*}
    Y_t := P_{\mathrm{VM}}(t),
\end{equation*}
where $P_{\mathrm{VM}}(t)$ is the VM-attributed CPU-package power
introduced above.

\begin{figure}[!ht]
 \includegraphics[width=\columnwidth]{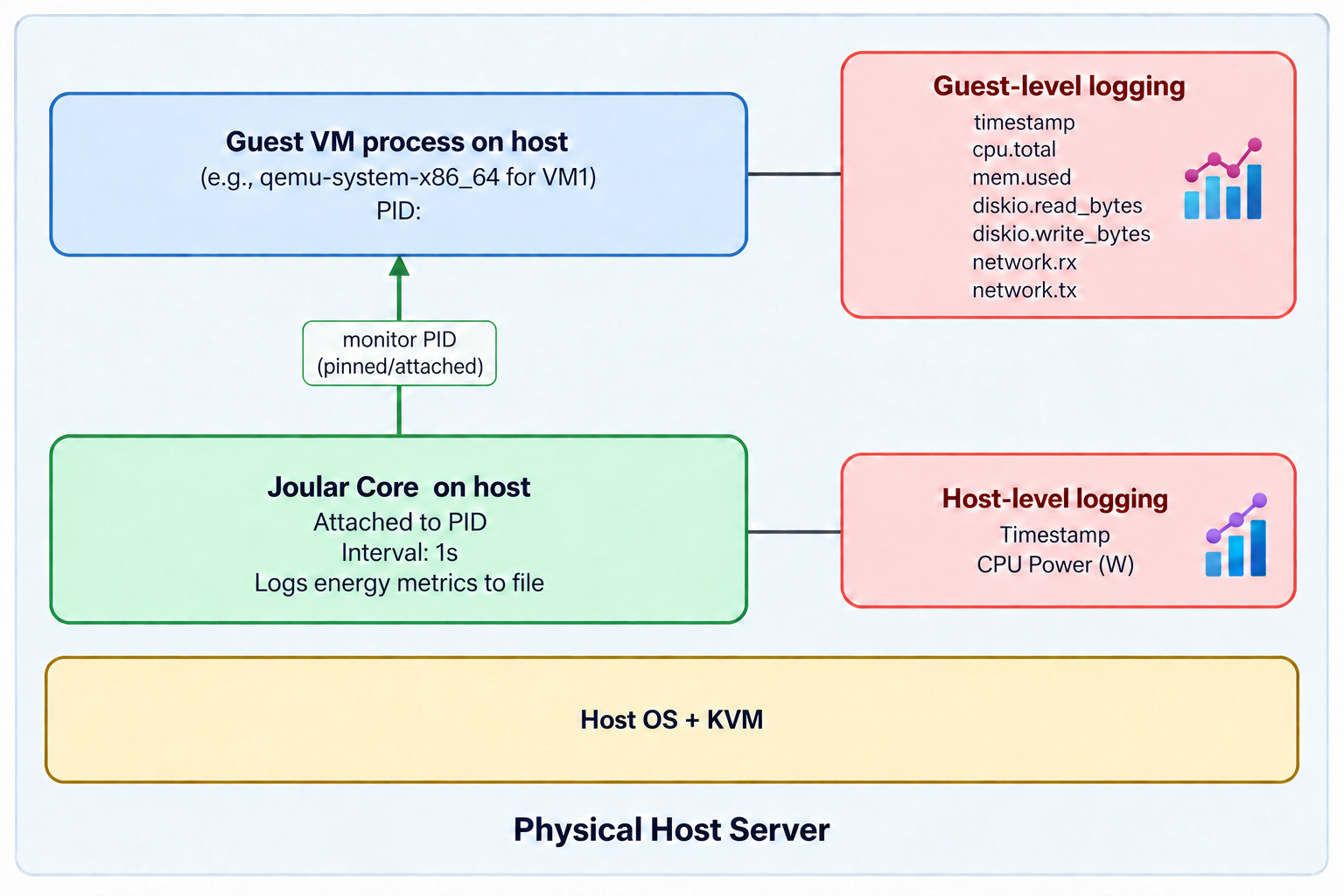}
 \caption{Overview of the dual-logging setup: RAPL-based power logging on the host and resource utilization logging on the guest.}
 \label{fig:datacapturingsetup}
\end{figure}

The learning task is to infer the mapping from guest-observable metrics
\(\mathbf{x}_t\) to VM-attributed power \(Y_t\) using synchronized guest-side
measurements and host-side power targets. After training, predictions are given by
\begin{equation*}
    \widehat{Y}_t = f(\mathbf{x}_t),
\end{equation*}
where \(f\) is the learned regression function. The timestamps are not included as predictor variables in
$\mathbf{x}_t$. They are recorded only to synchronize the host-side
power measurements with the guest-side metric stream and to construct
temporally valid training and test partitions. Figure~\ref{fig:datacapturingsetup} summarizes this dual-logging setup, with
RAPL-based VM-attributed power recorded on the host and the corresponding
resource and performance metrics collected inside the guest VM.

\subsection{Guest-Side Metrics}
All predictor variables are collected from within the guest VM at
approximately 1~s intervals. High-level operating-system metrics are recorded
with Glances \cite{glances}, including CPU utilization, memory utilization, disk I/O,
network I/O, system load, and process counts. To complement these high-level metrics, we collect low-level
performance counters using \textit{perf} \cite{linuxperf}  at 1~s intervals.
Counters common to the Intel and AMD configurations include CPU cycles,
instructions, block-I/O request issue and completion events, and
network transmit and receive events. Where supported, reference cycles
are also recorded. Architecture-specific memory-activity counters are
collected to characterize memory traffic: on Intel, these include
LLC-miss demand reads, retired LLC-miss loads, and retired load
instructions; on AMD, load/store dispatch events and DRAM/memory-fill
events are used instead. The EPYC 7443P exposes separate local and
remote memory-fill events rather than the aggregate DRAM event used on
the other AMD systems.

Thus, the collected guest-side variables span the main resource
dimensions relevant to VM power consumption: CPU activity, memory
usage and memory traffic, storage I/O, network I/O, and system load.
Only guest-observable variables are used as predictor variables, i.e. features for the model to be trained; no host-side
power or RAPL measurements are included in $\mathbf{x}_t$.

Guest-side perf accesses virtualized performance counters under guest OS access controls (e.g., CAP\_PERFMON or root privileges), without requiring direct access to physical host power interfaces. Counter availability depends on hypervisor configuration and cloud-provider restrictions; where unavailable, estimation remains possible using basic guest OS metrics (F1), but with reduced accuracy (Table S14 in the supplemental material). This preserves host–guest security isolation by avoiding direct exposure of physical host RAPL registers, thereby reducing the attack surface associated with privileged hardware interfaces.

\subsection{Timestamp Alignment and Data Preprocessing}
The host and guest data streams are joined along the timestamp dimension. Timestamps are therefore used for synchronization, not as predictive features.
Clock synchronization was provided by the default time-synchronization
services of the operating systems. All physical hosts ran Ubuntu
24.04.4 LTS with systemd-timesyncd, while all guest VMs ran
Ubuntu 26.04 LTS with Chrony. Both host and guest systems used UTC, and
no manual changes were made to the default synchronization
configuration.

Timestamps were converted to UTC and floored to one-second resolution.
Observations falling within the same one-second bin were averaged to a
single numeric value, and the guest- and host-side streams were then
inner-joined on the one-second timestamp. Invalid timestamps, non-numeric
measurements, and non-finite values were treated as missing and removed;
constant predictors were also excluded. No additional outlier filtering
or explicit warm-up/cool-down trimming was applied. Interval
perf counters (CPU cycles, instructions, reference cycles where
available, memory-activity counters, block-I/O events, and network
transmit/receive events) were converted to rates using the actual elapsed
sampling interval reported by perf.

\subsection{Prediction Models and Baselines}

We use a Gradient Boosting Regressor (GBR) as the model because it can represent nonlinear interactions among heterogeneous resource metrics while remaining inexpensive to train and evaluate. A lightweight inference model is particularly attractive for guest-side deployment because monitoring and prediction should add little resource overhead to the VM being measured.
We use the scikit-learn implementation of GBR, with 200 regression trees, learning rate of $0.1$, squared-error loss, and otherwise default settings. The choice of a conventional, instantaneous regressor is deliberate: the contribution lies in guest-side power estimation and its systematic evaluation, rather than a new prediction architecture. Temporal dependence is explicitly addressed through autocorrelation analysis and temporally blocked evaluation with exclusion gaps (Section~\ref{subsec:temporal-train-test-splitting}; Figs. S1–S3 in the supplemental materials), while directly exploiting temporal histories through sequence models is left for future work.

\subsection{Feature Sets and Ablation}
We consider $5$ feature sets in order to determine how
much predictive value is obtained from increasingly rich guest
side observability. We evaluate five guest-side feature sets, summarized in Table~\ref{tab:feature-sets}. Detailed feature definitions and platform-specific counter mappings are provided in Table~S4 and S5 in the supplemental material. Features unavailable on a given platform, or constant within a given
dataset, were excluded before model fitting. F5 denotes the union of the
available F1--F4 features. We performed the feature-set ablation on selected within-platform experiments from RQ1; all other experiments use the full available
feature set (F5). The results (Table~S14 in the supplemental
material) show that adding CPU-performance metrics (F2) improves
prediction over basic resource metrics (F1) in all seven evaluated
combinations. The full feature set (F5) achieves the highest mean $R^2$ in five of these seven combinations.
 
\begin{table}[t]
\centering
\caption{Guest-side feature sets used for model training.}
\label{tab:feature-sets}
\footnotesize
\setlength{\tabcolsep}{3pt}
\renewcommand{\arraystretch}{1.15}

\begin{tabularx}{\columnwidth}{
|>{\raggedright\arraybackslash}p{0.55cm}|
 >{\raggedright\arraybackslash}p{1.55cm}|
 >{\raggedright\arraybackslash}X|
}
\hline
\textbf{Set} & \textbf{Feature set} & \textbf{Features} \\
\hline
 
F1 &
Basic &
\makecell[l]{
cpu.total \\
mem.used \\
diskio.vda.read\_bytes\_rate\_per\_sec \\
diskio.vda.write\_bytes\_rate\_per\_sec \\
network.enp1s0.bytes\_recv\_rate\_per\_sec \\
network.enp1s0.bytes\_sent\_rate\_per\_sec
}
\\
\hline

F2 &
Basic + CPU performance &
\makecell[l]{
F1 + \\
perf.cycles\_per\_sec \\
perf.instructions\_per\_sec \\
perf.ref\_cycles\_per\_sec \\
perf.ipc
}
\\
\hline

F3 &
Basic + memory performance &
\makecell[l]{
F1 + \\
perf.dram\_read\_bytes\_est\_per\_sec \\
perf.l3\_load\_misses\_per\_sec \\
perf.loads\_per\_sec
}
\\
\hline

F4 &
Basic + I/O performance &
\makecell[l]{
F1 + \\
perf.block\_rq\_issue\_per\_sec \\
perf.block\_rq\_complete\_per\_sec \\
perf.net\_dev\_xmit\_per\_sec \\
perf.netif\_receive\_skb\_per\_sec
}
\\
\hline

F5 &
All &
Union of F1--F4
\\
\hline

\end{tabularx}
\end{table}

\section{Experimental Design}

The experimental design consists of a factorial core experiment supplemented by
targeted controlled experiments. The core design crosses hardware platform and
workload family, allowing the effects of workload and hardware variation, and
their interaction, to be examined systematically. Additional experiments isolate
concurrent workload composition, VM resource allocation, and monitoring
overhead. This structure provides an adequate coverage of the main deployment factors.

\subsection{Hardware Platforms}
\label{sec:hardware-platforms}
The study includes $6$ physical CPU platforms spanning Intel and AMD systems:

\begin{itemize}
    \item AMD EPYC 4245P, 6 cores;
    \item AMD EPYC 4345P, 8 cores;
    \item AMD EPYC 7443P, 24 cores;
    \item Intel Xeon E3-1270 v6, 4 cores; 
    \item Intel Xeon E-2286G, 6 cores;
    \item Intel Xeon 6527P, 24 cores.
\end{itemize}

The hardware platforms were selected subject to practical resource availability and cost constraints, while spanning a range of CPU generations and core counts. This diversity supports the experimental design by allowing the study to distinguish within-platform accuracy from cross-workload and cross-hardware generalization. In particular, the spread from small 4--8 core systems to larger 24 core server platforms provides a test of whether the learned guest-metric-to-power relationship remains stable across host architectures and scales. Detailed hardware platform specifications are provided in Table~S3 in the supplemental material.

\subsection{Virtual Machine Configuration}
The reference VM configuration used Ubuntu 24.04 as the operating system on the physical host servers, and x86-64 Ubuntu 26.04 LTS guest VMs under KVM/QEMU with a Q35 virtual-machine model, virtio-based storage and networking, and an ext4 root filesystem. Experiment-controlled quantities such as vCPU and RAM allocation are specified separately in the corresponding experimental-design sections. Detailed guest-VM specifications are provided in Table S6 in the supplemental material.

\subsection{Workloads}
To obtain resource-utilization patterns representative of common real-world VM workloads, we use four workload families chosen to exercise CPU, memory, network, and storage resources in different ways while remaining representative of practical server applications. Within each workload, randomness is introduced in load intensity, timing, concurrency, or task composition to generate a broad range of resource-utilization states over time.

\subsubsection{Linux Kernel Compilation}
Linux kernel compilation is used primarily as a CPU-intensive workload.
The workload builds Linux 7.1.6 for 3~hours using repeated compilation episodes. Each active episode lasts 45--180~s and randomly selects both the
build target (full kernel, drivers, filesystems, networking, or kernel
subtree) and the compilation parallelism, with \textit{make -j} levels
spanning from 1 up to the number of allocated vCPUs. The build tree is
initially cleaned and is cleaned again after a random 5--10 active episodes;
if a build completes early, cleaning is also used to sustain compilation
activity. To introduce lower-load periods, a 15--30~s idle interval is
inserted with 20\% probability after an active episode.

\subsubsection{PostgreSQL / pgbench}
A PostgreSQL workload is included to represent transactional database and
storage-I/O activity. We run pgbench for 3~hours as a sequence of
randomized load episodes drawn from five regimes---light, low, medium,
high, and burst---with probabilities 0.10, 0.20, 0.30, 0.30, and 0.10,
respectively. Episode durations range from 20--180~s depending on regime,
with client concurrency varying from 2--4 (light) up to 48--64 (burst);
worker-thread counts are selected according to the VM vCPU allocation.
The built-in pgbench profiles tpcb-like, simple-update, and
select-only are sampled with probabilities 0.50, 0.30, and 0.20,
respectively, and 15--30~s idle periods are inserted with 20\% probability.

\subsubsection{Apache Web Serving}
An Apache web server workload is used to represent common network- and request-processing activity.
The server exposes static HTML, Perl CGI, and PHP endpoints, requested by Apache Bench running on an external server, for 3~hour in randomized fixed-duration episodes. Five load regimes---light, low,
medium, high, and burst---are sampled with probabilities 0.10, 0.20, 0.30, 0.30,
and 0.10, with episode durations of 20--180~s and total request concurrency ranging from
3--9 up to 150--240 depending on regime. For each episode, concurrency is split
across static-heavy, CGI-heavy, PHP-heavy, or balanced endpoint profiles; the
three endpoint loads run concurrently. Idle intervals of 15--30~s are inserted with 20\% probability.

\subsubsection{iperf3}
An iperf3 workload is included as a pure network-throughput workload,
complementing the request-processing behavior of the Apache experiment.
Traffic is generated between the VM and an external server for 3~h in
randomized episodes spanning light, low, medium, high, and burst regimes
with probabilities 0.10, 0.20, 0.30, 0.30, and 0.10. Depending on regime,
aggregate target throughput ranges from 100~Mbit/s to 7~Gbit/s using
1--8 parallel streams and episode durations of 20--180~s. Upload,
download, and bidirectional modes are selected randomly, and 15--30~s
idle intervals are inserted with 20\% probability.

\subsection{Main Hardware and Workload Experiment}

The main experiment evaluates the same four workload families on five of the
six hardware platforms described in Section~\ref{sec:hardware-platforms}, using a common VM
configuration of 4 vCPUs and 8~GB RAM:
\begin{equation*}
    5\ \text{hardware platforms} \times 4\ \text{workloads}
    = 20\ \text{experimental cells}.
\end{equation*}

The AMD EPYC 4345P is reserved for the controlled VM-resource allocation experiment described in Section~\ref{subsec:vm-resource-var-exp}.

\subsection{Concurrent Workload Experiment}

Concurrent or mixed workloads are evaluated on the AMD EPYC 4245P using the
4-vCPU, 8-GB VM configuration. The workload pairs are:
\begin{itemize}
    \item kernel compilation + Apache;
    \item kernel compilation + iperf3;
    \item kernel compilation + pgbench;
    \item Apache + pgbench.
\end{itemize}

For each experiment, the two workload generators are run concurrently within
the same VM over the same 3-hour measurement period. Each workload
retains its independently randomized load schedule, so episode boundaries,
load intensities, and idle periods are not synchronized between the two
components. The resulting interference and overlap therefore vary naturally
throughout the run. 

\subsection{VM Resource Variation Experiment}
\label{subsec:vm-resource-var-exp}
Kernel compilation was selected for the VM-configuration experiment because it provides a sustained, repeatable, CPU-intensive workload with a direct connection to the RAPL-based CPU-package power target. It is also a practical server-side task whose parallelism can adapt naturally to different vCPU allocations. The same stochastic compilation procedure was retained across configurations, with the available \textit{make -j} levels adjusted to span from one thread up to the number of allocated vCPUs; no other workload-specific changes were introduced. Five VM resource configurations were used, varying vCPU allocation
between 2 and 8 and RAM allocation between 4 and 16 GB.
The configurations were (2, 8), (4, 4), (4, 8), (4, 16), and (8, 8),
where each pair denotes (vCPUs, RAM in GB).

\subsection{Monitoring-Overhead Experiment}
\label{sec:monitoring-overhead-experiment}
Monitoring overhead is evaluated on all six hardware platforms by running the kernel compilation workload with monitoring ON and OFF, respectively, in three separate runs. Each run uses a fixed 300-second measurement window and begins from a cleaned kernel tree.
In the OFF condition, only kernel compilation is executed; in the ON
condition, the same workload is executed while Glances and perf
collect the same guest-side metrics used in the main experiments at 1-second intervals.
Joular Core runs continuously on the physical host in both conditions.

Within each pair, the two conditions are separated by 30~s, with 60~s
between pairs. Condition order is alternated as OFF--ON, ON--OFF, and
OFF--ON to reduce bias. The primary
comparison is the paired difference in host-side power and energy during
the 300-s workload window; kernel-build progress is also recorded as a
workload-throughput indicator.

\subsection{Temporal Train–Test Splitting}
\label{subsec:temporal-train-test-splitting}
The guest VM metrics are sampled at one-second intervals and exhibit
substantial short-range autocorrelation. Consequently, randomly assigning
individual observations to training and test sets can place temporally
adjacent, strongly correlated observations in both sets. This can lead to memorization of observed patterns rather than genuine generalization, leading to overly optimistic estimates of prediction performance. We therefore use repeated temporally blocked train--test splitting
with exclusion gaps to separate training and test observations. For each
dataset, 20\% of the available observations are held out as one contiguous test
block. All observations within 120~s before and after this block are excluded
from training, giving the split structure:
\[
    \text{training}\;|\;120\text{-s gap}\;|\;
    \text{test block}\;|\;120\text{-s gap}\;|\;\text{training}.
\]
The remaining observations on either side form the training set. Test-block
locations are sampled uniformly from all positions that leave at least 50
training observations after application of the exclusion interval, and the
procedure is repeated 100 times.

The 120-s exclusion interval was selected from autocorrelation
analysis of the power target and guest-side predictors across workloads and hardware platforms; representative autocorrelation functions are shown in Figures S1--S3 in the supplemental material. Dependence varied between signals and experiments, but was predominantly concentrated at substantially shorter lags; 120~s was therefore adopted as a common conservative separation rather than
tuning the gap separately for each dataset. This protocol is used for all
train-test data splitting.

\subsection{Evaluation Metrics}

Prediction performance is evaluated using the coefficient of determination
($R^2$), mean absolute error (MAE), root mean squared error (RMSE), and
normalized mean absolute error (NMAE). $R^2$ measures the fraction of variation
in the observed power explained by the model; values close to one indicate
strong predictive agreement, while negative values indicate performance worse
than predicting the test-set mean. MAE gives the average absolute prediction
error in watts, whereas RMSE gives greater weight to large individual errors
and is therefore complementary to MAE.

To compare errors across workloads and hardware platforms with different
absolute power levels, we define
\begin{equation*}
    \mathrm{NMAE}
    =
    100\,
    \frac{\mathrm{MAE}}
         {\overline{Y}_{\mathrm{test}}}
    \%,
\end{equation*}
where $\overline{Y}_{\mathrm{test}}$ is the mean observed power in the
corresponding test set. Normalizing by the mean test power expresses the error
relative to the characteristic power level of the evaluated domain and therefore
permits more meaningful comparison across hosts and workloads than absolute
error alone.

For transfer experiments, degradation relative to the corresponding
within-domain result is quantified by
$
    \Delta R^2
    =
    R^2_{\mathrm{transfer}}
    -
    R^2_{\mathrm{within}}.
$
Thus, negative $\Delta R^2$ indicates loss of predictive performance under
transfer. Prediction bias, defined as the mean signed error
$\overline{\hat{Y}-Y}$, is additionally used as a diagnostic where systematic
over- or under-prediction is relevant.

For experiments using repeated temporal holdouts, all metrics are computed
separately for each of the 100 splits and summarized by their mean and standard
deviation. For experiments with a complete independent held-out domain, such as
leave-one-workload-out or leave-one-hardware-out evaluation, metrics are
computed directly over the full held-out dataset. Confidence intervals are used
where explicitly reported for paired overhead measurements.

\section{RQ1 --- Within-Platform Power Estimation}

Table~\ref{tab:rq1-within-platform} summarizes the within-platform prediction performance obtained with
the full feature set. Across the 20 hardware--workload combinations, prediction
accuracy is generally high, but varies noticeably with workload and platform.
Kernel compilation is the most consistently predictable workload, with
$R^2$ values between 0.952 and 0.985 and NMAE between 1.97\% and 5.84\%.
The Apache workload also performs strongly on most platforms, with $R^2$ between 0.887 and
0.959. PostgreSQL/pgbench shows greater variation, ranging from $R^2=0.786$
on the Xeon E-2286G to $0.935$ on the EPYC 7443P.

\begin{figure}[t]
\centering
\includegraphics[width=\columnwidth]{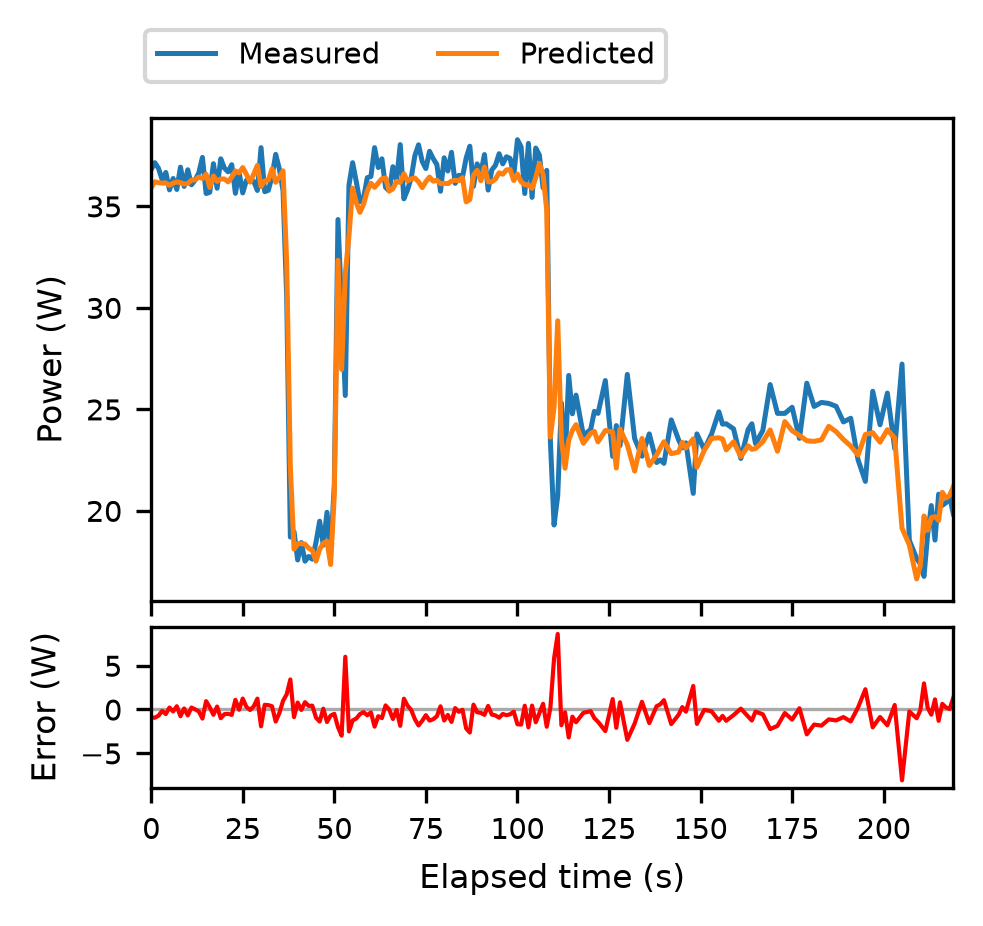}
\caption{Representative within-platform prediction for kernel compilation on
the Xeon 6527P. The upper panel compares measured and predicted VM-associated
power over a held-out interval; the lower panel shows the signed prediction
error, $\hat{Y}-Y$.}
\label{fig:rq1-timeseries}
\end{figure}

Figure~\ref{fig:rq1-timeseries} shows a representative temporal example of
within-domain prediction for kernel compilation on the Xeon 6527P. The model
tracks both the major changes in power level and most of the shorter-timescale
variation, with the largest residuals occurring around abrupt transitions.

The largest degradation occurs for iperf3, particularly on the Xeon
6527P ($R^2=0.734$) and in relative-error terms on the Xeon E-2286G
(NMAE $=22.38\%$). This indicates that within-platform prediction quality
depends not only on the host architecture but also on how strongly the workload
activity is coupled to the CPU-package power target. Overall, the results show
that guest-observable metrics can estimate VM-attributed power accurately
within a fixed hardware/workload domain, while network-dominated workloads are
less consistently predictable than CPU-intensive workloads.

\begin{table}[t]
\centering
\caption{Within-platform prediction performance using the full feature set.}
\label{tab:rq1-within-platform}
\small
\setlength{\tabcolsep}{3.0pt}
\renewcommand{\arraystretch}{1.08}
\begin{tabular}{l l r r r r}
\hline
\textbf{Platform} &
\textbf{Workload} &
\textbf{$\mathbf{R^2}$} &
\textbf{MAE} &
\textbf{RMSE} &
\textbf{NMAE} \\
&
&
&
\textbf{(W)} &
\textbf{(W)} &
\textbf{(\%)} \\
\hline

E3-1270 v6 & kernelcomp & 0.985 & 0.609 & 1.682 & 1.97 \\
E3-1270 v6 & apache & 0.936 & 2.688 & 3.877 & 10.30 \\
E3-1270 v6 & pgbench & 0.857 & 1.853 & 3.233 & 5.93 \\
E3-1270 v6 & iperf3 & 0.932 & 1.661 & 2.555 & 15.52 \\
E-2286G & kernelcomp & 0.961 & 2.450 & 3.900 & 5.58 \\
E-2286G & apache & 0.945 & 3.875 & 5.280 & 10.09 \\
E-2286G & pgbench & 0.786 & 4.550 & 6.530 & 8.47 \\
E-2286G & iperf3 & 0.916 & 2.495 & 4.510 & 22.38 \\
Xeon 6527P & kernelcomp & 0.952 & 1.231 & 1.922 & 5.07 \\
Xeon 6527P & apache & 0.887 & 1.248 & 2.014 & 6.94 \\
Xeon 6527P & pgbench & 0.920 & 1.329 & 1.846 & 6.09 \\
Xeon 6527P & iperf3 & 0.734 & 3.073 & 4.406 & 18.21 \\
Epyc 4245P & kernelcomp & 0.970 & 1.589 & 2.760 & 4.68 \\
Epyc 4245P & apache &  0.959 & 1.728 & 3.063 & 4.64 \\
Epyc 4245P & pgbench & 0.882 & 2.524 & 3.522 & 5.35 \\ 
Epyc 4245P & iperf3 & 0.855 & 2.123 & 2.916 & 12.81 \\
Epyc 7443P & kernelcomp & 0.967 & 1.268 & 1.985 & 5.68 \\
Epyc 7443P & apache & 0.936 & 0.969 & 1.828 & 4.52 \\
Epyc 7443P & pgbench & 0.935 & 0.949 & 1.660 & 2.92 \\
Epyc 7443P & iperf3 & 0.865 & 2.503 & 3.801 & 14.79 \\

\hline
\end{tabular}
\end{table}

\subsection{Feature-Importance Analysis}
To examine which guest-observable signals drive the predictions, we
analyzed the representative temporally blocked model for each of the
20 hardware--workload combinations using grouped permutation importance
\cite{gregorutti2015grouped}. This measures the importance of a feature
group by jointly permuting its variables in the held-out data and
recording the resulting decrease in prediction performance
($\Delta R^2$). Features were grouped into CPU performance, CPU
utilization, memory, storage I/O, and network I/O. We additionally
trained one workload-pooled model per hardware platform by combining
the four workload training portions while preserving the same temporal
holdout procedure. Within each model, positive grouped permutation
importances were normalized to sum to 100\%, allowing resource-family
contributions to be compared across platforms and workloads.

\begin{figure}[t]
    \centering
    \includegraphics[width=\columnwidth]
    {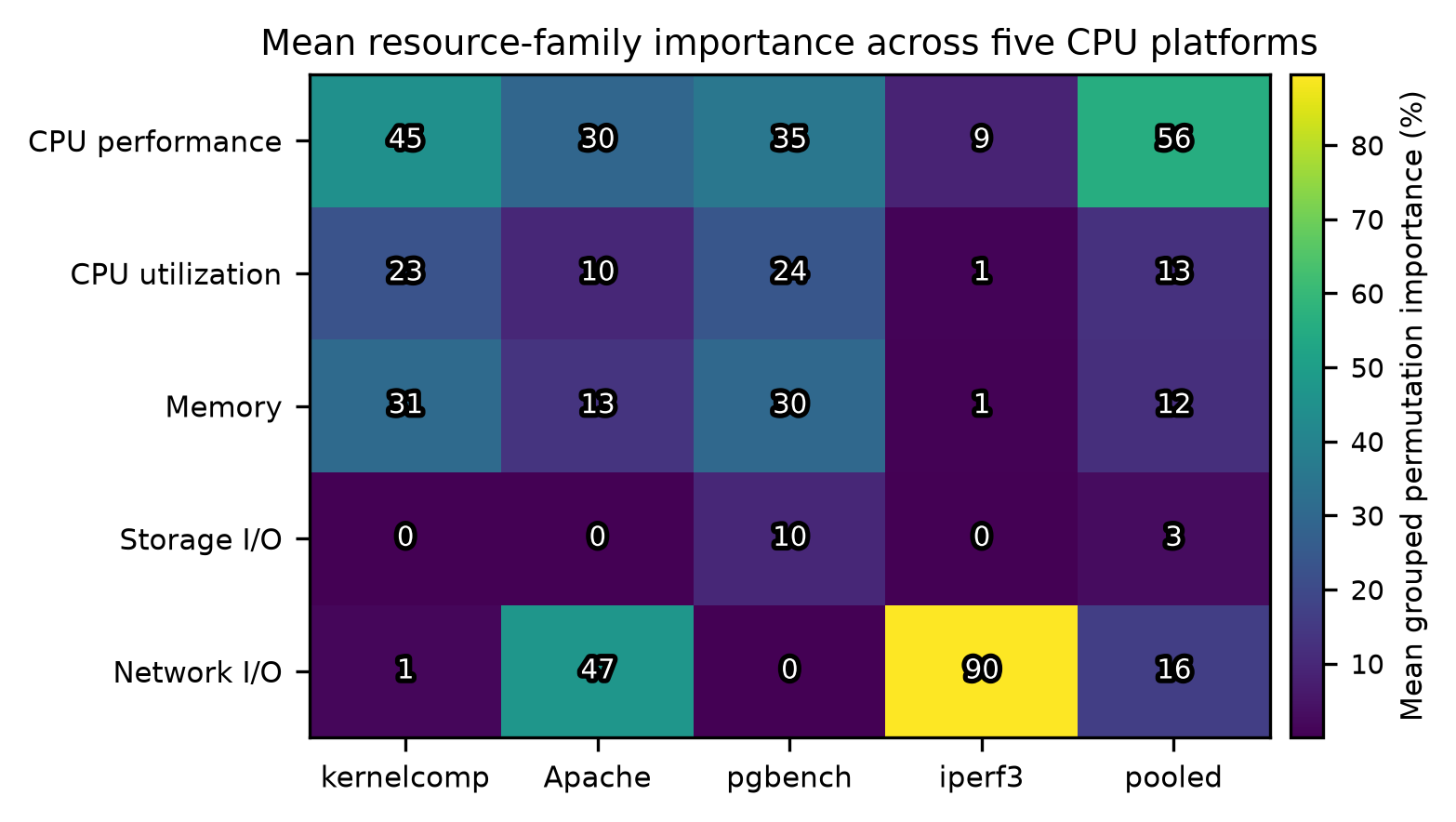}
    \caption{Mean grouped permutation-importance share across the five
    CPU platforms. The pooled column corresponds to the
    within-platform workload-pooled model on each hardware platform.}
    \label{fig:fi-mean}
\end{figure}

Figure~\ref{fig:fi-mean} shows the mean normalized importance across
the five main CPU platforms. Kernel compilation is dominated by CPU
performance (44.9\%) and memory-related signals (30.8\%), while
pgbench exhibits a broader CPU--memory--storage signature
(35.5\%, 30.4\%, and 9.9\%, respectively). Apache shows a mixed
network/CPU signature, with network I/O contributing 46.9\% on
average. For iperf3, network I/O is overwhelmingly dominant
(89.6\%), consistent with its network-throughput workload design.
The workload-pooled models rely primarily on CPU-performance signals
(56.0\%), with smaller contributions from network I/O (16.4\%),
CPU utilization (12.6\%), and memory (12.1\%).

\begin{figure}[t]
    \centering
    \includegraphics[width=\columnwidth]
    {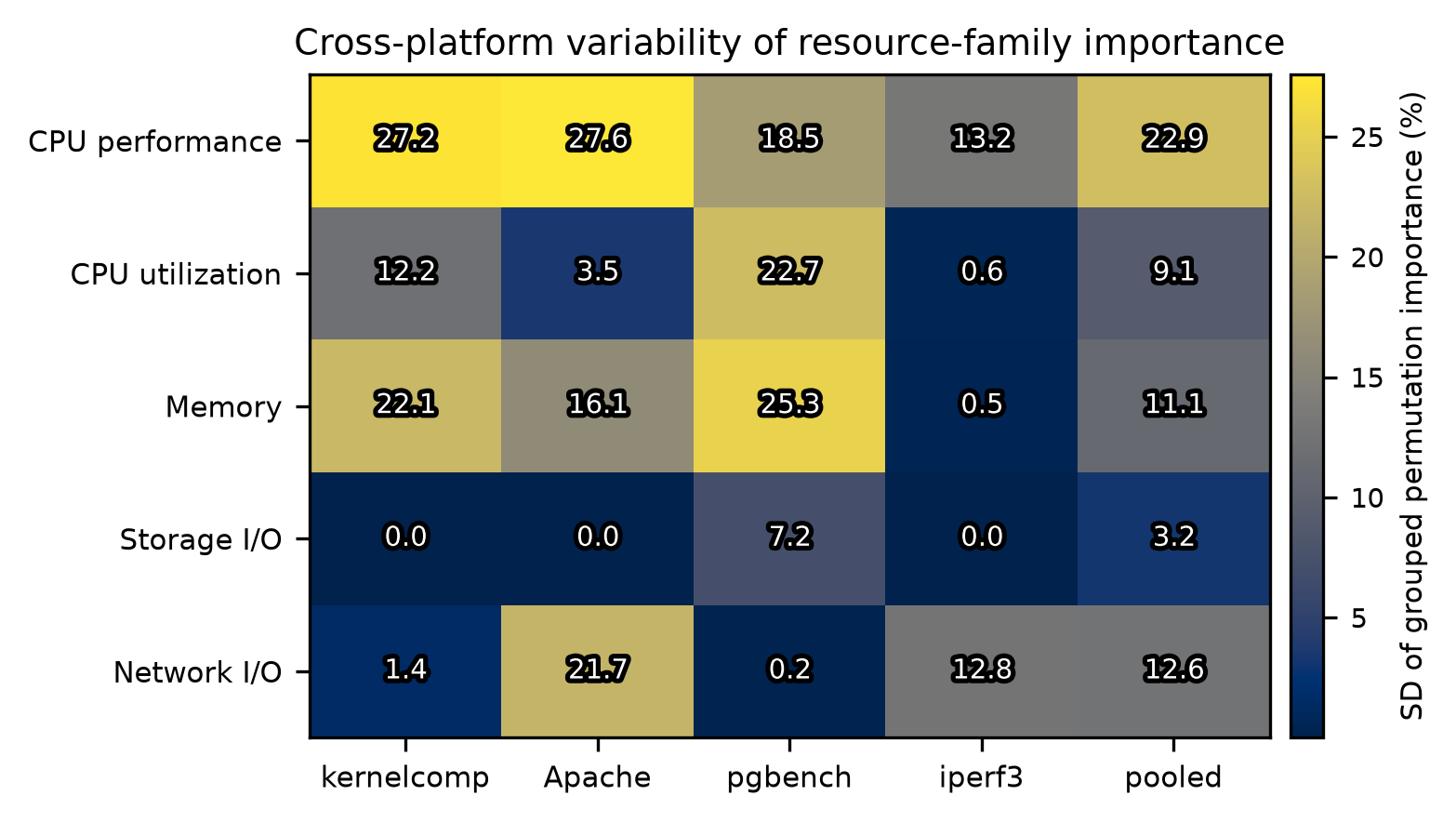}
    \caption{Cross-platform variability of grouped permutation
    importance. Each cell gives the standard deviation, in percentage
    points, of the normalized resource-family importance across the
    five CPU platforms.}
    \label{fig:fi-sd}
\end{figure}

Figure~\ref{fig:fi-sd} shows the corresponding cross-platform
standard deviation. The relative contribution of CPU-performance and
memory-related signals varies substantially across hardware for
kernel compilation and pgbench, while Apache also shows marked
variation between CPU and network contributions. In contrast,
iperf3 retains a stable qualitative signature: CPU-utilization and
memory importance are consistently negligible, although the relative
weight assigned to network and CPU-performance signals varies across
platforms. The pooled models likewise retain some hardware dependence,
particularly in CPU-performance importance (SD 22.9 percentage
points). Thus, workload identity determines the broad resource
signature, while hardware influences how importance is distributed
within that signature.

As a robustness check, permutation- and impurity-based feature
rankings were strongly concordant across platforms, with mean
Spearman rank correlations of 0.906, 0.863, 0.838, 0.757, and 0.838
for kernelcomp, Apache, pgbench, iperf3, and the workload-pooled
models, respectively. The lower agreement for iperf3 is consistent
with several correlated network-related predictors sharing similar
predictive information.

More details are provided in the supplemental material: Figures~S4--S8 show the platform-specific grouped permutation-importance results, Table~S7 summarizes agreement between permutation- and impurity-based feature rankings, Tables~S8--S12 report the top-five feature-level permutation rankings, and Figures~S9--S13 provide complementary SHAP summaries for the workload-pooled models.

\section{RQ2 --- Workload Generalization}

To evaluate generalization to unseen workloads, we use a
leave-one-workload-out (LOWO) design. For each hardware platform,
the model is trained on data from three workload families and
evaluated on the fourth, previously unseen workload. This procedure
is repeated for each workload family and hardware platform,
providing a direct assessment of cross-workload transfer without
training data from the target workload.

Tables~III and IV summarize the resulting prediction performance
and its change relative to the corresponding within-workload
results from RQ1. Transfer performance varies substantially across
both workloads and hardware platforms. Figure~\ref{fig:lowo-kernelcomp} illustrates leave-one-workload-out transfer to kernel
compilation on the Xeon 6527P. The model captures the main
power transitions but systematically underpredicts several
higher-power regimes, indicating that it learns some transferable
workload characteristics without fully reproducing the target
workload's power levels.

\begin{table}[t]
\centering
\caption{Leave-one-workload-out \(R^2\) for each hardware platform and held-out workload.}
\label{tab:rq2-lowo-r2}
\small
\setlength{\tabcolsep}{3.0pt}
\renewcommand{\arraystretch}{1.08}
\begin{tabular}{lrrrr}
\hline
\textbf{Platform} &
\textbf{kernelcomp} &
\textbf{Apache} &
\textbf{pgbench} &
\textbf{iperf3} \\
\hline

E3-1270 v6 & 0.914 & 0.630 & 0.493 & 0.130 \\
E-2286G    & 0.442 & 0.661 & -1.153 & -2.351 \\
Xeon 6527P & 0.756 & 0.738 & 0.751 & 0.208 \\
EPYC 4245P & 0.709 & 0.558 & 0.515 & -1.395 \\
EPYC 7443P & 0.053 & 0.153 & -1.331 & 0.516 \\

\hline
\end{tabular}
\end{table}

\begin{table}[t]
\centering
\caption{Change in \(R^2\) under leave-one-workload-out generalization, relative to the corresponding within-workload \(R^2\) from RQ1: \(\Delta R^2 = R^2_{\mathrm{LOWO}} - R^2_{\mathrm{within}}\).}
\label{tab:rq2-lowo-delta}
\small
\setlength{\tabcolsep}{3.0pt}
\renewcommand{\arraystretch}{1.08}
\begin{tabular}{lrrrr}
\hline
\textbf{Platform} &
\textbf{kernelcomp} &
\textbf{Apache} &
\textbf{pgbench} &
\textbf{iperf3} \\
\hline

E3-1270 v6 & -0.071 & -0.306 & -0.364 & -0.802 \\
E-2286G    & -0.519 & -0.284 & -1.939 & -3.267 \\
Xeon 6527P & -0.196 & -0.149 & -0.169 & -0.526 \\
EPYC 4245P & -0.261 & -0.401 & -0.367 & -2.250 \\
EPYC 7443P & -0.914 & -0.783 & -2.266 & -0.349 \\

\hline
\end{tabular}
\end{table}

\begin{center}
\includegraphics[width=\columnwidth]{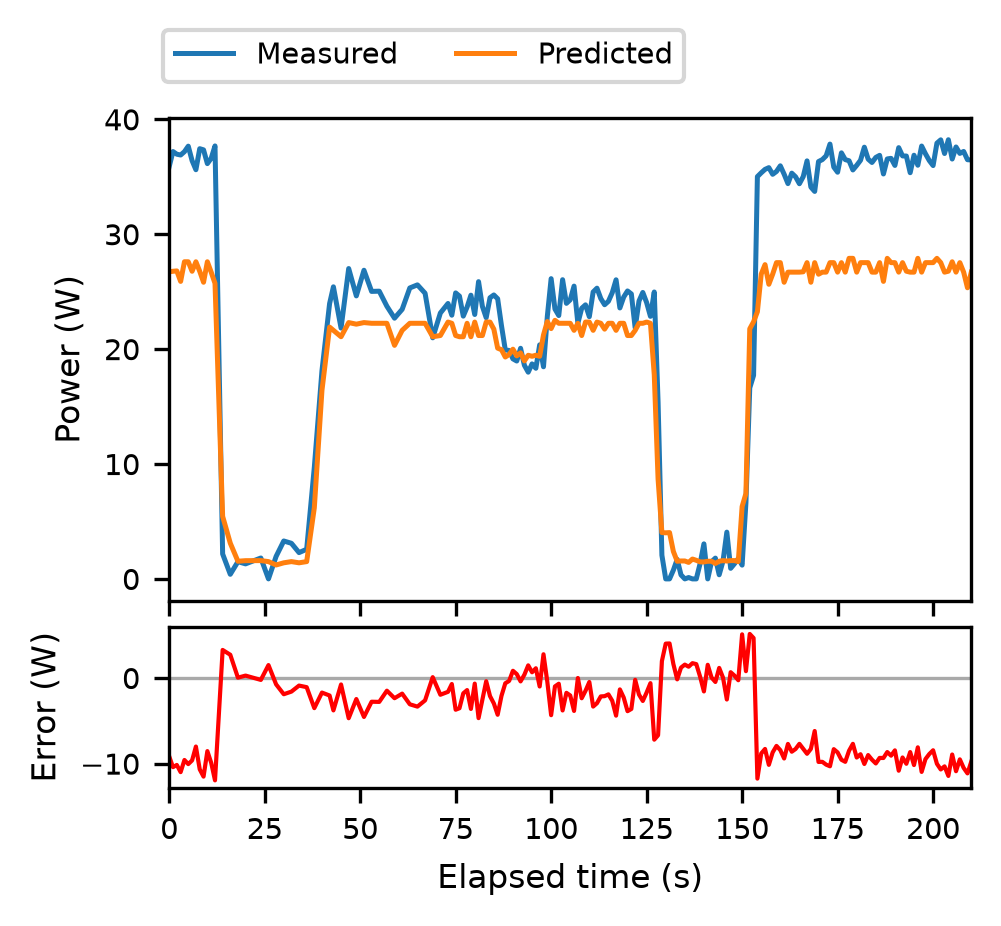}
\captionof{figure}{Leave-one-workload-out prediction for held-out
kernel compilation on the Xeon 6527P. The transferred model captures the
principal temporal transitions but shows systematic underprediction in several
higher-power regimes. The lower panel shows signed prediction error,
$\hat{Y}-Y$.}
\label{fig:lowo-kernelcomp}
\end{center}

Figure~\ref{fig:lowo-iperf3} illustrates a weaker transfer case
on the same hardware. For held-out iperf3, the prediction is markedly flatter,
misses a large transient, and does not fully reproduce the later reduction in
power level. This shows how poor workload generalization can arise from both
missed short-lived events and systematic errors.

\begin{center}
\includegraphics[width=\columnwidth]{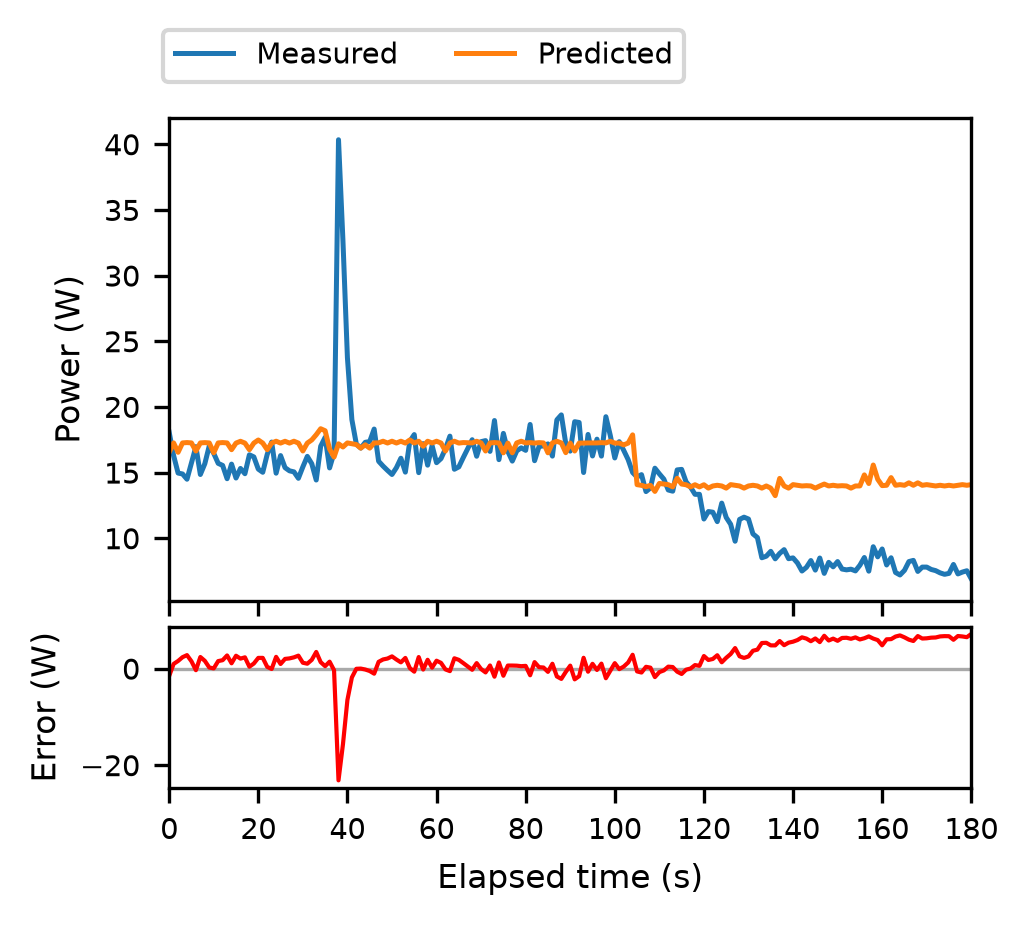}
\captionof{figure}{Leave-one-workload-out prediction for held-out iperf3 on
the Xeon 6527P. In this weaker-transfer case, the model misses a large
transient and under-represents later changes in power level, illustrating
behavior not apparent from aggregate error metrics alone. The lower panel
shows signed prediction error, $\hat{Y}-Y$.}
\label{fig:lowo-iperf3}
\end{center}

\section{RQ3 --- Hardware Generalization}

We evaluate hardware generalization in two complementary settings. First, we
measure zero-shot transfer to an unseen hardware platform using a
leave-one-hardware-out (LOHO) design. Second, we evaluate whether a single
multi-platform model can learn a common mapping across heterogeneous hardware
when all target platforms are represented during training.

\subsection{Leave-One-Hardware-Out Generalization}
  
For each held-out hardware platform, the model is trained on the remaining
four main-experiment platforms and evaluated on the complete set of observations
from the unseen platform. All four workloads are included in both the pooled training data and the held-out
platform test data. Because the target hardware platform is entirely absent from
training, this experiment measures zero-shot hardware transfer rather than
within-platform interpolation.

Table~\ref{tab:rq3-loho} summarizes the LOHO results. Generalization to unseen
hardware is substantially more variable than the within-domain performance
reported in RQ1. The strongest transfer is obtained for the EPYC 4245P, with
$R^2=0.774$ and NMAE $=22.88\%$, followed by the EPYC 7443P
($R^2=0.625$, NMAE $=25.61\%$) and the Xeon E3-1270 v6
($R^2=0.566$, NMAE $=31.62\%$). Transfer to the Xeon E-2286G is weaker
($R^2=0.468$, NMAE $=40.51\%$), while the Xeon 6527P exhibits negative
overall $R^2$ ($R^2=-0.373$), indicating performance below a mean-value
predictor on the held-out platform.

The per-workload LOHO results, provided in Table~S13 in the supplemental material, further show that hardware transfer is not
uniformly difficult across workloads. For example, the EPYC 4245P
retains high generalization performance for kernel compilation
($R^2=0.927$), whereas the Xeon 6527P shows severe degradation
for Apache and pgbench ($R^2=-2.399$ and $-2.890$, respectively).
Conversely, iperf3 generalizes relatively well on the EPYC 7443P
($R^2=0.838$). These results indicate that zero-shot hardware transfer
depends jointly on the target platform and workload characteristics,
and that strong performance on known hardware does not guarantee
reliable transfer to a previously unseen host.

\begin{table}[t]
\centering
\caption{Leave-one-hardware-out performance on each unseen hardware platform.}
\label{tab:rq3-loho}
\small
\setlength{\tabcolsep}{2.0pt}
\renewcommand{\arraystretch}{1.08}
\begin{tabular}{p{2.1cm}rrrr}
\hline
\textbf{Held-out platform} &
\textbf{$\mathbf{R^2}$} &
\textbf{MAE (W)} &
\textbf{RMSE (W)} &
\textbf{NMAE (\%)} \\
\hline
E3-1270 v6 & 0.566 & 7.614 & 10.087 & 31.62 \\
E-2286G    & 0.468 & 14.526 & 18.151 & 40.51 \\
Xeon 6527P & -0.373 & 7.737 & 10.914 & 39.53 \\
EPYC 4245P & 0.774 & 7.254 & 9.401 & 22.88 \\
EPYC 7443P & 0.625 & 5.761 & 7.114 & 25.61 \\
\hline
\end{tabular}
\end{table}

\subsection{Pooled Multi-Platform Model}
We pool observations from five hardware platforms and four workloads into a common training and test design. The objective is to test whether one pooled model can represent heterogeneous hardware when
all target platforms are represented during training, and therefore
complement the zero-shot LOHO setting. 

For each hardware--workload time series, the same temporally blocked evaluation
strategy used in RQ1 is applied, including the 120-s exclusion gap around the
test block. The training portions from all 20 hardware--workload combinations are
then concatenated to form a single pooled training set, and the corresponding
held-out portions are concatenated to form a pooled test set.

A single model is trained on the pooled training set and evaluated both globally
and separately for each hardware platform and workload. Table~\ref{tab:rq3-pooled} summarizes the pooled multi-platform results.

When all five hardware platforms are represented during training, the single
pooled model achieves an overall $R^2$ of $0.890\pm0.006$, MAE of
$3.858\pm0.109$~W, RMSE of $5.835\pm0.151$~W, and NMAE of
$13.87\pm0.45\%$ across the 100 temporally blocked evaluations. The small
variation across repetitions indicates that the result is insensitive to the
particular temporal blocks selected for testing.

\begin{figure}[t]
\centering
\includegraphics[width=\columnwidth]{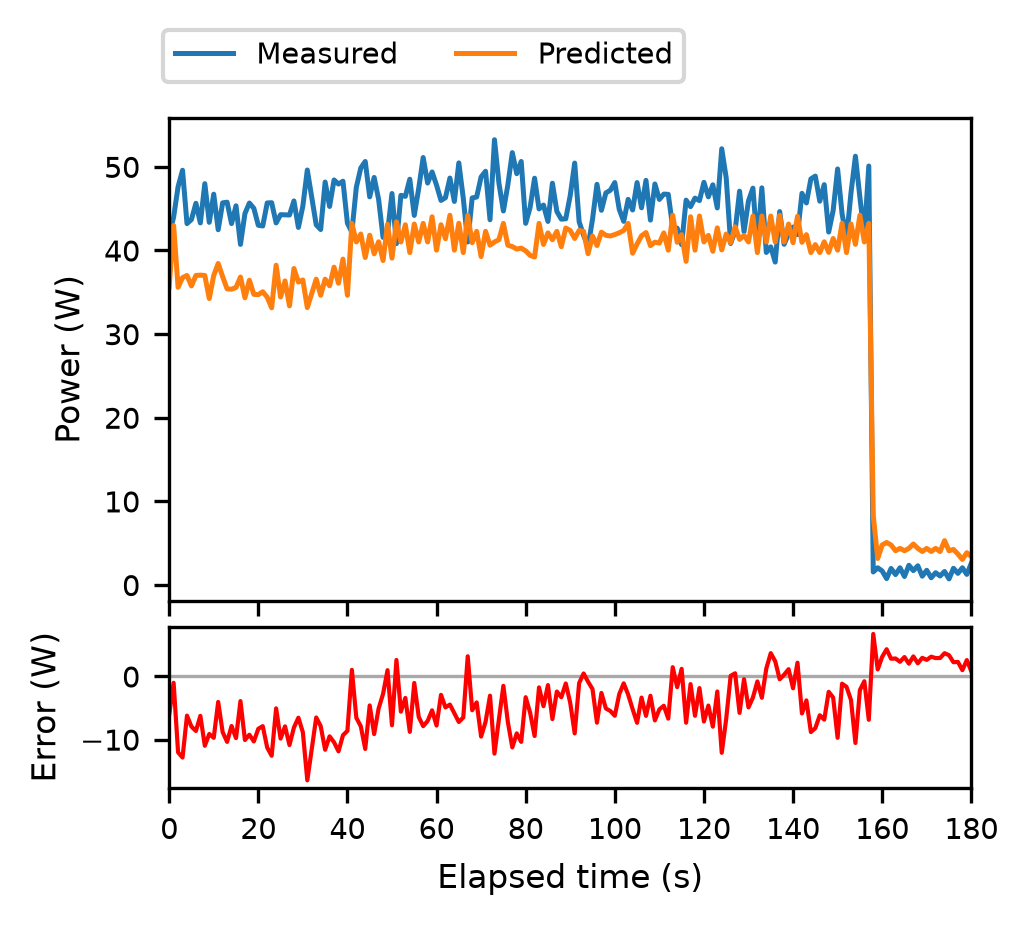}
\caption{Representative held-out prediction from the pooled multi-platform
model. The upper panel compares measured and predicted VM-associated power;
the lower panel shows the signed prediction error, $\hat{Y}-Y$.}
\label{fig:pooled-timeseries}
\end{figure}

Figure~\ref{fig:pooled-timeseries} shows a representative held-out
time-series segment from the pooled multi-platform model. The prediction
tracks the dominant temporal structure and major changes in power level,
while the residuals reveal the remaining systematic and short-timescale
errors that are not visible from aggregate metrics alone.

Performance remains strong on four of the five individual platforms. The
highest platform-level performance is obtained on the EPYC 4245P
($R^2=0.952\pm0.009$), followed by the E3-1270 v6
($0.889\pm0.026$), E-2286G ($0.862\pm0.012$), and EPYC 7443P
($0.856\pm0.017$). The Xeon 6527P remains more difficult, with
$R^2=0.546\pm0.064$, although its performance is substantially better than
the zero-shot LOHO transfer.

The contrast between the two RQ3 experiments is pronounced. Under LOHO,
$R^2$ ranges from $-0.373$ to $0.774$ across unseen platforms, whereas the
pooled model yields positive $R^2$ on every platform, ranging from $0.546$ to
$0.952$. For example, the Xeon 6527P improves from $R^2=-0.373$ when held
out entirely to $R^2=0.546$ when observations from that platform are included
in the pooled training data. Similar improvements occur for the remaining
platforms.

These results indicate that the principal difficulty is not the use of a
single model across heterogeneous hardware, but extrapolation to
hardware that is completely absent from training. A common multi-platform
model can therefore provide strong prediction performance when representative
training observations from the deployed hardware platforms are available,
whereas zero-shot transfer to previously unseen hardware remains less reliable.

\begin{table}[t]
\centering
\caption{Pooled multi-platform prediction performance, averaged over
100 temporally blocked evaluations.}
\label{tab:rq3-pooled}
\small
\setlength{\tabcolsep}{3.0pt}
\renewcommand{\arraystretch}{1.08}
\begin{tabular}{lrrrr}
\hline
\textbf{Platform} &
\textbf{$\mathbf{R^2}$} &
\textbf{MAE (W)} &
\textbf{RMSE (W)} &
\textbf{NMAE (\%)} \\
\hline
All platforms & 0.890 & 3.858 & 5.835 & 13.87 \\
E3-1270 v6    & 0.889 & 3.496 & 4.966 & 14.16 \\
E-2286G       & 0.862 & 6.495 & 9.027 & 17.68 \\
Xeon 6527P    & 0.546 & 3.833 & 5.776 & 18.71 \\
EPYC 4245P    & 0.952 & 2.675 & 4.000 & 8.02 \\
EPYC 7443P    & 0.856 & 2.970 & 4.147 & 12.70 \\
\hline
\end{tabular}
\end{table}

\section{RQ4 --- Generalization to Concurrent Workloads}

RQ4 differs from the pooled multi-platform experiment of RQ3 in that workload
composition is changed during data acquisition rather than only during model
training. In RQ3, observations collected separately across hardware platforms
and workloads are pooled for model training, whereas the RQ4 datasets
are generated by two workloads executing simultaneously within the
same VM. The measurements therefore include
contention and workload-interaction effects that are absent from the
corresponding single-workload runs. We consider four selected workload combinations (Table~\ref{tab:rq4-concurrent}) to examine a range of
concurrent workload interactions, rather than exhaustively testing all
possible combinations, which would substantially increase experimental
time and cost.

\subsection{Within-Concurrent Prediction}

We first evaluate each concurrent workload combination independently using the
same repeated temporally blocked procedure as in RQ1. For each mixture, a
contiguous 20\% test block is held out with a 120-s exclusion gap on both sides,
and the experiment is repeated 100 times. This establishes the prediction
performance achievable when representative concurrent observations are
available during training. 

Within-concurrent prediction is strong across all four workload combinations, see Table~\ref{tab:rq4-concurrent}.
Kernelcomp+iperf3 and kernelcomp+Apache achieve mean $R^2$ values of 0.959
and 0.951, respectively, while Apache+pgbench reaches 0.859.
Kernelcomp+pgbench is less consistent, with mean $R^2=0.756$. Despite this variation in $R^2$, within-concurrent prediction errors
remain small, with NMAE ranging from 2.16\% to 4.04\%, compared
with 3.70\% to 11.12\% under transfer. Concurrent execution is
therefore not intrinsically difficult to model when concurrent
observations are represented during training.

\subsection{Single-Workload to Concurrent-Workload Transfer}

We next test whether models trained only on the two constituent workloads
executing separately can generalize to their simultaneous execution. For each
mixture $A+B$, the corresponding single-workload datasets are concatenated for
training. A single transfer model is fitted for each workload pair and evaluated on the
same 100 concurrent test blocks used for the within-concurrent experiment.
This paired evaluation ensures that the within-concurrent and transfer metrics
are computed on identical target observations.

Table~\ref{tab:rq4-concurrent} summarizes the results. Transfer performance is
strong for kernelcomp+Apache and kernelcomp+iperf3, with mean $R^2$ values of
0.845 and 0.802, respectively. Kernelcomp+pgbench shows weaker transfer
($R^2=0.487$), while Apache+pgbench exhibits substantial degradation, with
mean $R^2=-0.178$ despite within-concurrent $R^2=0.859$. The corresponding
change in $R^2$ ranges from $-0.107$ for kernelcomp+Apache to $-1.037$ for
Apache+pgbench.

\begin{table}[t]
\centering
\caption{Concurrent-workload prediction and transfer performance. Within and
transfer metrics are averaged over the same 100 temporally blocked test
intervals.}
\label{tab:rq4-concurrent}
\small
\setlength{\tabcolsep}{1.5pt}
\renewcommand{\arraystretch}{1.08}
\begin{tabular}{p{2.7cm}rrrr}
\hline
\textbf{Workload mix} & 
\textbf{$\mathbf{R^2_{\rm within}}$} &
\textbf{$\mathbf{R^2_{\rm transfer}}$} &
\textbf{$\mathbf{\Delta R^2}$} &
\textbf{$\mathbf{NMAE}_{\mathrm{transfer}}$ (\%)} \\
\hline
Apache+pgbench
& 0.859 & -0.178 & -1.037 & 8.77 \\

kernelcomp+Apache
& 0.951 & 0.845 & -0.107 & 6.66 \\

kernelcomp+iperf3
& 0.959 & 0.802 & -0.157 & 11.12 \\

kernelcomp+pgbench
& 0.756 & 0.487 & -0.269 & 3.70 \\
\hline
\end{tabular}
\end{table}

Overall, concurrent workload states remain highly predictable when
represented during training, whereas transfer from isolated workloads is
workload-pair dependent. The degradation observed for some mixtures
indicates that simultaneous execution can alter the resource--power
relationship beyond what is captured by the constituent workloads alone.

\section{RQ5 --- Sensitivity to VM Resource Allocation}

RQ5 evaluates whether the learned guest-metrics-to-power relationship remains
stable when the VM resource allocation changes. Here, all experiments use the AMD
EPYC 4345P and the kernel-compilation workload. Five configurations are
considered: $(2,8)$, $(4,4)$, $(4,8)$, $(4,16)$, and $(8,8)$, where the
entries denote vCPU count and RAM in GB. The $(4,8)$ configuration is used as
the reference setting.

\subsection{Within-Configuration Prediction}

We first evaluate each configuration independently using the same 100-repeat
temporally blocked procedure as in RQ1. Prediction performance is consistently
high across all five configurations, see Table~\ref{tab:rq5-config}, where mean $R^2$ ranges from 0.935 to
0.973 and NMAE from 3.34\% to 4.45\%. This indicates that each allocation
regime is itself readily learnable when representative observations from that
configuration are available during training.

\subsection{Transfer from the Reference Configuration}

We next train a single model on the reference configuration
$(4,8)$ and evaluate it on the same 100 target test blocks used for the
within-configuration analysis. Transfer performance varies substantially with
the direction and magnitude of the allocation change. The strongest transfer is
observed for $(4,16)$, with $R^2=0.918$ and
$\Delta R^2=-0.047$. Transfer to $(8,8)$ remains reasonably strong
($R^2=0.816$), whereas performance decreases more substantially for
$(4,4)$ ($R^2=0.747$) and especially for $(2,8)$
($R^2=0.453$).

The corresponding NMAE values are 6.59\%, 17.36\%, 20.03\%, and 18.36\%
for $(4,16)$, $(8,8)$, $(4,4)$, and $(2,8)$, respectively. These results
show that zero-shot transfer across VM allocation changes is asymmetric and can
involve substantial calibration shifts even when the target configuration is
individually highly predictable.

\subsection{Pooled Multi-Configuration Model}

Finally, we train a single model on temporally separated training portions from
all five resource configurations and evaluate it on the corresponding held-out
test portions. The pooled model achieves an overall $R^2$ of 0.949, MAE of
1.872 W, RMSE of 2.988 W, and NMAE of 5.60\%.

Configuration-specific performance also remains strong, with $R^2$ values of
0.916, 0.932, 0.960, 0.943, and 0.957 for $(2,8)$, $(4,4)$, $(4,8)$,
$(4,16)$, and $(8,8)$, respectively. Table~\ref{tab:rq5-config} summarizes
the three evaluation settings.

Overall, the results indicate that VM resource allocation changes do not
prevent the use of a common predictive model when the relevant configurations
are represented during training. In contrast, zero-shot transfer from a single
reference configuration is less reliable, particularly for some reductions in
allocated resources. This suggests that practical deployment across varying VM
sizes benefits from multi-configuration training or calibration rather than
assuming a configuration-invariant mapping learned from one reference setting.
\begin{table}[t]
\centering
\caption{Sensitivity to VM resource allocation on the EPYC 4345P.
Within-configuration and pooled-model results are averaged over 100
temporally blocked evaluations. Transfer uses the 4-vCPU, 8-GB
configuration as the reference.}
\label{tab:rq5-config}
\small
\setlength{\tabcolsep}{2.8pt}
\renewcommand{\arraystretch}{1.08}
\begin{tabular}{lrrrr}
\hline
\textbf{Configuration} &
\textbf{$\mathbf{R^2_{\rm within}}$} &
\textbf{$\mathbf{R^2_{\rm transfer}}$} &
\textbf{$\mathbf{\Delta R^2}$} &
\textbf{$\mathbf{R^2_{\rm pooled}}$} \\
\hline
2 vCPU, 8 GB
& 0.935 & 0.453 & -0.482 & 0.916 \\

4 vCPU, 4 GB
& 0.969 & 0.747 & -0.223 & 0.932 \\

4 vCPU, 8 GB
& 0.973 & -- & -- & 0.960 \\

4 vCPU, 16 GB
& 0.965 & 0.918 & -0.047 & 0.943 \\

8 vCPU, 8 GB
& 0.964 & 0.816 & -0.148 & 0.957 \\
\hline
\end{tabular}
\end{table}

\section{RQ6 --- Monitoring and Inference Overhead}

Monitoring overhead is evaluated using three matched ON/OFF pairs on each of
the six hardware platforms using the kernel compilation workload. OFF means that VM metric
collection is disabled, whereas ON denotes that metric collection using Glances and perf is
enabled. Each run is analyzed over an identical 300-s interval. For a measured
quantity $X$, paired overhead is defined as
$
    \Delta X = X_{\mathrm{ON}} - X_{\mathrm{OFF}},
$
with relative overhead
\begin{equation*}
    100\frac{\Delta X}{X_{\mathrm{OFF}}}\%.
\end{equation*}

Table~\ref{tab:rq6-overhead} summarizes the results. We report changes in mean CPU power $\Delta P$ (W) and energy
consumption $\Delta E$ (Wh). Workload progress is approximated by
counting kernel-build actions (e.g., CC, LD, AR). Progress (\%) is the relative change in this count, while
E/action (\%) is the relative change in CPU energy per counted action. This is to account for the effect of monitoring on
workload efficiency.

The direct effect
of metric collection on host CPU power is small. Mean differences range from
$-0.744$ W to $+0.326$ W across the six platforms, corresponding to relative
changes between approximately $-0.97\%$ and $+0.49\%$. The associated
300-s energy differences are similarly small, ranging from $-0.062$ Wh to
$+0.027$ Wh. With only three matched pairs per platform, the confidence
intervals for the CPU-power differences include zero on all platforms.

The small negative mean differences observed on the EPYC 4345P and EPYC 7443P
should not be interpreted as monitoring-induced power savings. Their magnitude
is comparable to run-to-run variability, while the monitored runs also show
reduced kernel-compilation progress. Thus, measurement jitter and partial
displacement of useful workload activity by the monitoring processes are both
plausible contributors to the observed sign.

The effect on workload progress is more apparent than the direct power effect.
Kernel-compilation progress decreases by approximately 1--6\% across the
platforms. Consequently, energy consumed per unit of useful build progress
increases on all six platforms, by approximately 1.45--6.34\%. This indicates
that the principal monitoring cost is not a large increase in instantaneous
CPU power, but a modest loss of workload efficiency as monitoring processes
compete for VM resources.

Joular Core Process Power provides a consistent secondary result. Mean paired
changes remain small across the six platforms, ranging from approximately
$-1.28\%$ to $+0.84\%$, and do not reveal a substantially larger hidden
VM-attributed power overhead.

\begin{table}[t]
\centering
\caption{Monitoring overhead for Glances and perf on the six hardware
platforms. Values are means over three matched ON/OFF pairs, with each run
evaluated over the same 300-s interval.}
\label{tab:rq6-overhead}
\small
\setlength{\tabcolsep}{2.8pt}
\renewcommand{\arraystretch}{1.08}
\begin{tabular}{p{1.3cm}cccc}
\hline
\textbf{Platform} &
\textbf{$\mathbf{\Delta P}$ (W)} &
\textbf{$\mathbf{\Delta E}$ (Wh)} &
\textbf{Progress (\%)} &
\textbf{$E$/action (\%)} \\
\hline
E3-1270 v6 &  0.071 &  0.0059 & -4.00 & 4.41 \\
\arrayrulecolor{gray!35}\hline
E-2286G    &  0.097 &  0.0080 & -4.23 & 4.55 \\
\arrayrulecolor{gray!35}\hline
Xeon 6527P &  0.319 &  0.0266 & -1.01 & 1.45 \\
\arrayrulecolor{gray!35}\hline
EPYC 4245P &  0.326 &  0.0272 & -1.10 & 1.60 \\
\arrayrulecolor{gray!35}\hline
EPYC 4345P & -0.744 & -0.0620 & -2.66 & 1.74 \\
\arrayrulecolor{gray!35}\hline
EPYC 7443P & -0.367 & -0.0306 & -6.25 & 6.34 \\
\hline
\end{tabular}
\end{table} 

Inference overhead was evaluated separately on the Intel Xeon
E3-1270 v6 using three matched inference-on/off pairs under
kernel compilation (300\,s per condition), with Glances and
perf enabled throughout. The deployed F5 GBR operated
at 1~s intervals, with a median prediction latency of
1.25\,ms and near-complete inference coverage. Enabling
inference increased mean VM-attributed power by only
0.099\,W (0.231\%; 95\% CI: $[-0.143,\,0.604]$\%).
The estimated interval lies within the predefined
$\pm1$\% practical-equivalence margin, indicating negligible
additional inference power overhead under the tested
conditions, see Tables~S15 and S16 in the supplemental materials.

\section{Discussion}

The results demonstrate that guest-observable system metrics can support
accurate VM-associated power estimation when the deployment conditions
are represented during training. Within fixed hardware--workload domains
(RQ1), prediction is strong across most of the 20 combinations. Kernel
compilation is generally the most predictable workload, while the
network-dominated iperf3 workload exhibits the largest and most variable
errors. Prediction difficulty therefore depends on workload resource
behavior as well as hardware characteristics.

Workload generalization (RQ2) reveals that accurate within-domain
prediction does not guarantee transfer to unseen workload families.
Leave-one-workload-out performance varies substantially across workloads
and hardware platforms, with several combinations producing negative
$R^2$. The same workload can generalize well on one platform and poorly
on another, indicating that workload behavior and hardware jointly
influence transfer performance.

Hardware generalization (RQ3) distinguishes extrapolation to unseen
hardware from prediction across represented platforms. Zero-shot transfer
is inconsistent, including negative $R^2$ on the Xeon 6527P. In contrast,
a single pooled model achieves overall $R^2=0.890$ when all five hardware
platforms are represented during training. Heterogeneous hardware
therefore does not necessarily require separate models for each host,
but reliable prediction benefits from calibration data covering the
relevant platforms.

Concurrent execution (RQ4) introduces workload interactions that cannot
always be captured by pooling observations from separately executed
workloads. Concurrent combinations remain highly predictable when
represented during training, but transfer from isolated workloads
depends on the particular mixture. Kernelcomp+Apache and
kernelcomp+iperf3 retain strong transfer performance, whereas
Apache+pgbench degrades substantially. These differences indicate that
contention and simultaneous resource activity can alter the
guest-metric-to-power relationship, making representative concurrent
observations valuable for deployment.

VM resource allocation (RQ5) has a similar effect on generalization.
All five vCPU/RAM configurations are individually highly predictable,
but transfer from the reference $(4,8)$ configuration is asymmetric and
degrades particularly for some reduced-resource settings. A pooled
multi-configuration model achieves overall $R^2=0.949$, demonstrating
that a common model can accommodate different VM sizes when the
relevant allocation regimes are represented during training.
Multi-configuration calibration is therefore preferable to assuming
that a mapping learned at one VM size remains invariant.

Monitoring overhead (RQ6) introduces little direct CPU-power cost but
reduces useful workload progress. Enabling Glances and perf changes mean
host CPU power by less than approximately 1\% across the six platforms,
with confidence intervals for the paired differences including zero.
Kernel-compilation progress decreases by approximately 1--6\%, increasing
energy per unit of completed work by 1.45--6.34\%. The negative mean
power differences observed on the EPYC 4345P and 7443P do not establish
power savings; run-to-run variability and displacement of useful work
by monitoring are plausible explanations. The main runtime trade-off
is therefore reduced workload efficiency rather than substantially
increased instantaneous power consumption.

These findings favor a deployment model based on representative
calibration data covering the expected hardware platforms, workload
families, concurrent execution patterns, and VM resource allocations.
Such coverage allows a lightweight nonlinear estimator to operate
without exposing privileged host power interfaces to the guest.
The practical trade-off lies between the breadth of calibration data,
the desired deployment generality, and the measurable cost of collecting
guest-observable metrics.

\section{Conclusion and Future Work}

This study demonstrates that resource-utilization and performance
metrics observable within a virtual server can support accurate
estimation of its VM-attributed CPU-package power without requiring
privileged host-side measurements during inference. The systematic
evaluation across heterogeneous hardware, workloads, concurrent
execution, and VM resource configurations establishes that prediction
reliability depends primarily on the correspondence between training
data and deployment conditions.

A common model can achieve strong performance across multiple hardware
platforms and VM configurations when representative observations are
available during training, whereas zero-shot generalization to unseen
conditions is less reliable. The findings therefore support broad,
representative calibration rather than requiring separate models for
every deployment setting. Guest-side monitoring adds little direct
CPU-power overhead, although it introduces a modest reduction in
workload efficiency.

The present work opens several interesting directions for further
research and practical development. Deep learning and sequence models
could exploit temporal dependencies in resource utilization and
potentially achieve higher prediction accuracy than the tree-based
regression models considered here. 
Other interesting extensions include additional virtualization
technologies, such as Xen, VMware, and Hyper-V, containerized
environments such as Docker, and operating systems beyond Linux, particularly Windows. Hardware accelerators, especially
GPUs, are another natural extension. Lightweight adaptation and
calibration methods could further improve generalization to unseen
hardware and workloads using limited additional data.  Finally, integrating such estimators into practical monitoring tools
and cloud platforms could enable service providers to offer tenants
direct visibility into their virtual servers' estimated power and
energy consumption, based on actual resource activity rather than
coarse proportional attribution of host power, supporting more
accurate and fine-grained energy-aware operation and management.


\bibliographystyle{IEEEtran}
\bibliography{refs}

\clearpage 
\onecolumn
\newpage
\begin{center} \LARGE\bfseries Supplementary Material \end{center}
\renewcommand{\thetable}{S\arabic{table}}
\renewcommand{\thefigure}{S\arabic{figure}}
\setcounter{figure}{0} 
\setcounter{table}{0}

\vspace{0.5em}

\begin{table}[!htbp]
\centering
\caption{Comparison with representative prior approaches to server and
VM-level power estimation.}
\label{tab:related-comparison}
\small
\begin{tabular}{p{2.4cm} p{2.4cm} p{2.0cm} p{2.2cm} p{2.0cm} p{2.1cm} p{2.1cm}}
\hline
Study &
Prediction target &
Guest-only inference &
Runtime host / hypervisor access &
Hardware diversity &
Transfer / generalization &
Concurrent / VM-config. evaluation \\
\hline

Joulemeter [10] &
VM power &
No &
Yes &
Multiple production servers &
Not primary focus &
No \\

VMeter [12] &
VM and host power &
No &
Yes &
Single experimental platform &
Multiple workloads &
No \\

CloudMonitor [13] &
Server power &
No &
Calibration / provider-side &
Hardware-specific calibration &
Limited &
No \\

Povoa et al. [14] &
Whole-system power &
No &
OS/system access &
Multiple hardware configurations &
Cross-hardware modeling &
No \\ 

Veni and Bhanu [16] &
VM power &
No &
Performance-counter access &
Limited &
Multiple workloads &
No \\

Present work &
VM-associated CPU-package power &
Yes, after training &
Training only &
Six Intel/AMD platforms &
Workload and hardware transfer &
Yes \\
\hline
\end{tabular}
\end{table}

\begin{table}[!htbp]
\centering
\caption{Input features and reported prediction performance
of representative prior approaches. Results are reproduced
from the respective studies and are not directly comparable
because prediction targets, hardware, workloads, and
evaluation protocols differ.}
\label{tab:related-accuracy}
\small
\renewcommand{\arraystretch}{1.3}
\begin{tabular}{p{2.5cm} p{5.2cm} p{7.0cm}}
\hline
Study & Input features / families & Reported prediction performance \\
\hline

Joulemeter [10] &
CPU utilization and power states, memory activity,
disk I/O; optional cache counters &
Approximately 1--3 W error in reported multi-VM
power-estimation experiments. \\

VMeter [12] &
CPU, cache, disk, and DRAM resource-activity
counters &
93\% mean and 94\% median prediction accuracy
against external power measurements. \\

CloudMonitor [13] &
CPU, memory, disk, and network utilization &
3.91\% average prediction error on the evaluated
server configuration. \\

Povoa et al. [14] &
OS resource counters for CPU, memory,
disk I/O, and network activity &
Best reported result: 99.94\% accuracy and 
6.32 W error. \\

Veni and Bhanu [16] &
Hardware PMCs (CPU cycles, DRAM accesses,
instruction-cache fetches, LLC misses);
disk I/O statistics &
91--92\% reported prediction accuracy  
using nonlinear SVR across benchmark workloads. \\

Present work &
Guest CPU, memory, disk and network metrics;
virtualized CPU and I/O performance counters &
Pooled multi-platform:
$R^2=0.890$, MAE $=3.858$ W,
NMAE $=13.87\%$;
best within-platform:
$R^2=0.985$, MAE $=0.609$ W,
NMAE $=1.97\%$. \\

\hline
\end{tabular}
\end{table}

\begin{table*}[t]
\centering
\caption{Hardware and software characteristics of the physical host platforms.}
\label{tab:supp-hardware}
\small
\setlength{\tabcolsep}{4pt}
\renewcommand{\arraystretch}{1.15}

\begin{tabularx}{\textwidth}{
    >{\raggedright\arraybackslash}p{2.8cm}
    >{\raggedright\arraybackslash}X
    >{\raggedright\arraybackslash}X
    >{\raggedright\arraybackslash}X
    >{\raggedright\arraybackslash}X
    >{\raggedright\arraybackslash}X
    >{\raggedright\arraybackslash}X
}
\toprule
\textbf{Attribute}
&
\textbf{Intel Xeon E3-1270 v6}
&
\textbf{Intel Xeon E-2286G}
&
\textbf{Intel Xeon 6527P}
&
\textbf{AMD EPYC 4245P}
&
\textbf{AMD EPYC 4345P}
&
\textbf{AMD EPYC 7443P}
\\
\midrule

Microarchitecture / generation
& Kaby Lake / Xeon E3-1200 v6
& Coffee Lake / Xeon E-2200
& Granite Rapids-SP / Xeon 6
& Zen 5 / EPYC 4005
& Zen 5 / EPYC 4005
& Zen 3 / EPYC 7003 (Milan)
\\

Sockets
& 1
& 1
& 1
& 1
& 1
& 1
\\

Cores / threads
& 4 / 8
& 6 / 12
& 24 / 48
& 6 / 12
& 8 / 16
& 24 / 48
\\

Base / max frequency
& 3.8 / 4.2 GHz
& 4.0 / 4.9 GHz
& 3.0 / 4.2 GHz
& 3.9 / 5.484 GHz
& 3.8 / 5.581 GHz
& 2.8 GHz
\\

Host memory
& 32 GB DDR4-2400 ECC, 2$\times$16 GB Micron
& 32 GB DDR4-2667 ECC, 2$\times$16 GB Micron
& 256 GB DDR5-6400 ECC, 8$\times$32 GB Micron
& 32 GB DDR5-5600 ECC, 2$\times$16 GB Micron
& 64 GB DDR5-5200 ECC, 2$\times$32 GB Micron
& 256 GB DDR4-3200 ECC, 8$\times$32 GB
\\

Storage
& 2$\times$240 GB SATA SSD
& 2$\times$960 GB SATA SSD
& 2$\times$1.92 TB NVMe SSD
& 2$\times$960 GB NVMe SSD
& 2$\times$1.92 TB NVMe SSD
& 2$\times$480 GB SAS SSD
\\

Host OS
& Ubuntu 24.04.4 LTS
& Ubuntu 24.04.4 LTS
& Ubuntu 24.04.4 LTS
& Ubuntu 24.04.4 LTS
& Ubuntu 24.04.4 LTS
& Ubuntu 24.04.4 LTS
\\

Kernel
& 6.8.0-138-generic
& 6.8.0-138-generic
& 6.8.0-139-generic
& 6.8.0-138-generic
& 6.8.0-138-generic
& 6.8.0-138-generic
\\

Virtualization stack
& KVM / QEMU / libvirt
& KVM / QEMU / libvirt
& KVM / QEMU / libvirt
& KVM / QEMU / libvirt
& KVM / QEMU / libvirt
& KVM / QEMU / libvirt
\\

RAPL domains
& package-0, core, dram
& package-0, core, dram
& package-0, dram
& package-0, core
& package-0, core
& package-0, core
\\

Joular Core version
& 0.1.0
& 0.1.0
& 0.1.0
& 0.1.0
& 0.1.0
& 0.1.0
\\

Host NIC / driver
& ixgbe, 10 Gb/s
& ixgbe, 10 Gb/s
& mlx5\_core, 25 Gb/s
& ice, 25 Gb/s
& ice, 25 Gb/s
& bnxt\_en, 25 Gb/s
\\

VM NIC model
& virtio
& virtio
& virtio
& virtio
& virtio
& virtio
\\

\bottomrule
\end{tabularx}
\end{table*}


\begin{table*}[t]
\centering
\caption{Guest-side model features and their definitions. Rate-based
perf features are normalized by the actual elapsed sampling
interval.}
\label{tab:supp-features}
\footnotesize
\setlength{\tabcolsep}{4pt}
\renewcommand{\arraystretch}{1.15}

\begin{tabularx}{\textwidth}{
|>{\raggedright\arraybackslash}p{0.8cm}|
 >{\raggedright\arraybackslash}p{4.4cm}|
 >{\raggedright\arraybackslash}p{1.5cm}|
 >{\raggedright\arraybackslash}X|
 >{\raggedright\arraybackslash}p{1.5cm}|
}
\hline
\textbf{Set} &
\textbf{Model feature} &
\textbf{Source} &
\textbf{Meaning / construction} &
\textbf{Unit}
\\
\hline

F1 &
\path|cpu.total| &
Glances &
Total CPU utilization observed inside the guest VM. &
\%
\\
\hline

F1 &
\path|mem.used| &
Glances &
Guest memory currently in use. &
bytes
\\
\hline

F1 &
\path|diskio.vda.read_bytes_rate_per_sec| &
Glances &
Read throughput of the guest virtual block device vda. &
bytes/s
\\
\hline

F1 &
\path|diskio.vda.write_bytes_rate_per_sec| &
Glances &
Write throughput of the guest virtual block device vda. &
bytes/s
\\
\hline

F1 &
\path|network.enp1s0.bytes_recv_rate_per_sec| &
Glances &
Guest network receive throughput on interface enp1s0. &
bytes/s
\\
\hline

F1 &
\path|network.enp1s0.bytes_sent_rate_per_sec| &
Glances &
Guest network transmit throughput on interface enp1s0. &
bytes/s
\\
\hline

F2 &
\path|perf.cycles_per_sec| &
perf &
CPU cycles divided by the measured perf sampling interval. &
events/s
\\
\hline

F2 &
\path|perf.instructions_per_sec| &
perf &
Instructions divided by the measured perf sampling interval. &
events/s
\\
\hline

F2 &
\path|perf.ref_cycles_per_sec| &
perf &
Reference CPU cycles divided by the measured perf sampling interval.
Unavailable on the EPYC 7443P collection. &
events/s
\\
\hline

F2 &
\path|perf.ipc| &
derived &
Instructions divided by CPU cycles. &
ratio
\\
\hline

F4 &
\path|perf.block_rq_issue_per_sec| &
perf &
Linux block-layer request-issue events divided by the measured
sampling interval. &
events/s
\\
\hline

F4 &
\path|perf.block_rq_complete_per_sec| &
perf &
Linux block-layer request-completion events divided by the measured
sampling interval. &
events/s
\\
\hline

F4 &
\path|perf.net_dev_xmit_per_sec| &
perf &
Linux network-device transmit events divided by the measured
sampling interval. &
events/s
\\
\hline

F4 &
\path|perf.netif_receive_skb_per_sec| &
perf &
Linux receive-SKB events divided by the measured sampling interval. &
events/s
\\
\hline

\end{tabularx}
\end{table*}

\begin{table*}[t]
\centering
\caption{Architecture-specific memory-performance features used in
feature set F3. Intel and AMD PMU events are retained with distinct
semantics rather than treated as equivalent hardware counters.}
\label{tab:supp-memory-features}
\footnotesize
\setlength{\tabcolsep}{4pt}
\renewcommand{\arraystretch}{1.15}

\begin{tabularx}{\textwidth}{
|>{\raggedright\arraybackslash}p{2.0cm}|
 >{\raggedright\arraybackslash}p{4.1cm}|
 >{\raggedright\arraybackslash}p{4.3cm}|
 >{\raggedright\arraybackslash}X|
}
\hline
\textbf{Platform} &
\textbf{Derived model feature} &
\textbf{Raw perf event(s)} &
\textbf{Construction / interpretation}
\\
\hline

Intel &
\path|perf.dram_read_bytes_est_per_sec| &
\path|offcore_requests.l3_miss_demand_data_rd| &
Demand-data reads that miss the LLC are converted to a rate and
multiplied by 64 bytes per cache line to obtain an estimated DRAM-read
traffic rate.
\\
\hline

Intel &
\path|perf.l3_load_misses_per_sec| &
\path|mem_load_retired.l3_miss| &
Retired load instructions missing the L3 cache, divided by the
measured perf interval.
\\
\hline

Intel &
\path|perf.loads_per_sec| &
\path|mem_inst_retired.all_loads| &
Retired load instructions divided by the measured perf interval.
\\
\hline

AMD EPYC 4245P / 4345P &
\path|perf.sys_dram_io_fills_per_sec| &
\path|ls_dmnd_fills_from_sys.dram_io_all| &
System demand fills serviced through DRAM I/O, divided by the measured
perf interval.
\\
\hline

AMD EPYC 7443P &
\path|perf.sys_dram_io_fills_per_sec| &
\path|ls_dmnd_fills_from_sys.mem_io_local| +
\path|ls_dmnd_fills_from_sys.mem_io_remote| &
Local and remote system-memory demand fills are summed and divided by
the measured perf interval to form the same AMD derived feature.
\\
\hline

AMD &
\path|perf.loads_dispatched_per_sec| &
\path|ls_dispatch.ld_dispatch| &
Dispatched load operations divided by the measured perf interval.
\\
\hline

AMD &
\path|perf.stores_dispatched_per_sec| &
\path|ls_dispatch.store_dispatch| &
Dispatched store operations divided by the measured perf interval.
\\
\hline

\end{tabularx}
\end{table*}

\begin{table}[t]
\centering
\caption{Common guest-VM software and virtual-hardware configuration.}
\label{tab:supp-guest-config}
\small
\setlength{\tabcolsep}{4pt}
\renewcommand{\arraystretch}{1.15}

\begin{tabularx}{\columnwidth}{
|>{\raggedright\arraybackslash}p{2.7cm}|
 >{\raggedright\arraybackslash}X|
}
\hline
\textbf{Attribute} & \textbf{Configuration} \\
\hline

Guest OS &
Ubuntu 26.04 LTS
\\
\hline

Kernel &
Linux 7.0.0 
\\
\hline

Architecture &
x86-64
\\
\hline

Virtualization &
KVM
\\
\hline

Virtual hardware model &
QEMU Q35 / ICH9
\\
\hline

Virtual disk &
100 GB
\\
\hline

Disk type &
virtio block device
\\
\hline

Root filesystem &
ext4
\\
\hline

VM NIC model &
virtio
\\
\hline

Network MTU &
1500 bytes
\\
\hline

Glances version &
4.3.3
\\
\hline

perf version &
7.0.12/7.0.14
\\
\hline

\end{tabularx}
\end{table}

\clearpage 

\begin{figure}[H]
    \centering
    \caption{Autocorrelation ($\rho$) of the VM-attributed power target at 1~Hz sampling.
    The dashed horizontal lines indicate $|\rho|=0.1$.}
    \includegraphics[width=0.95\textwidth]{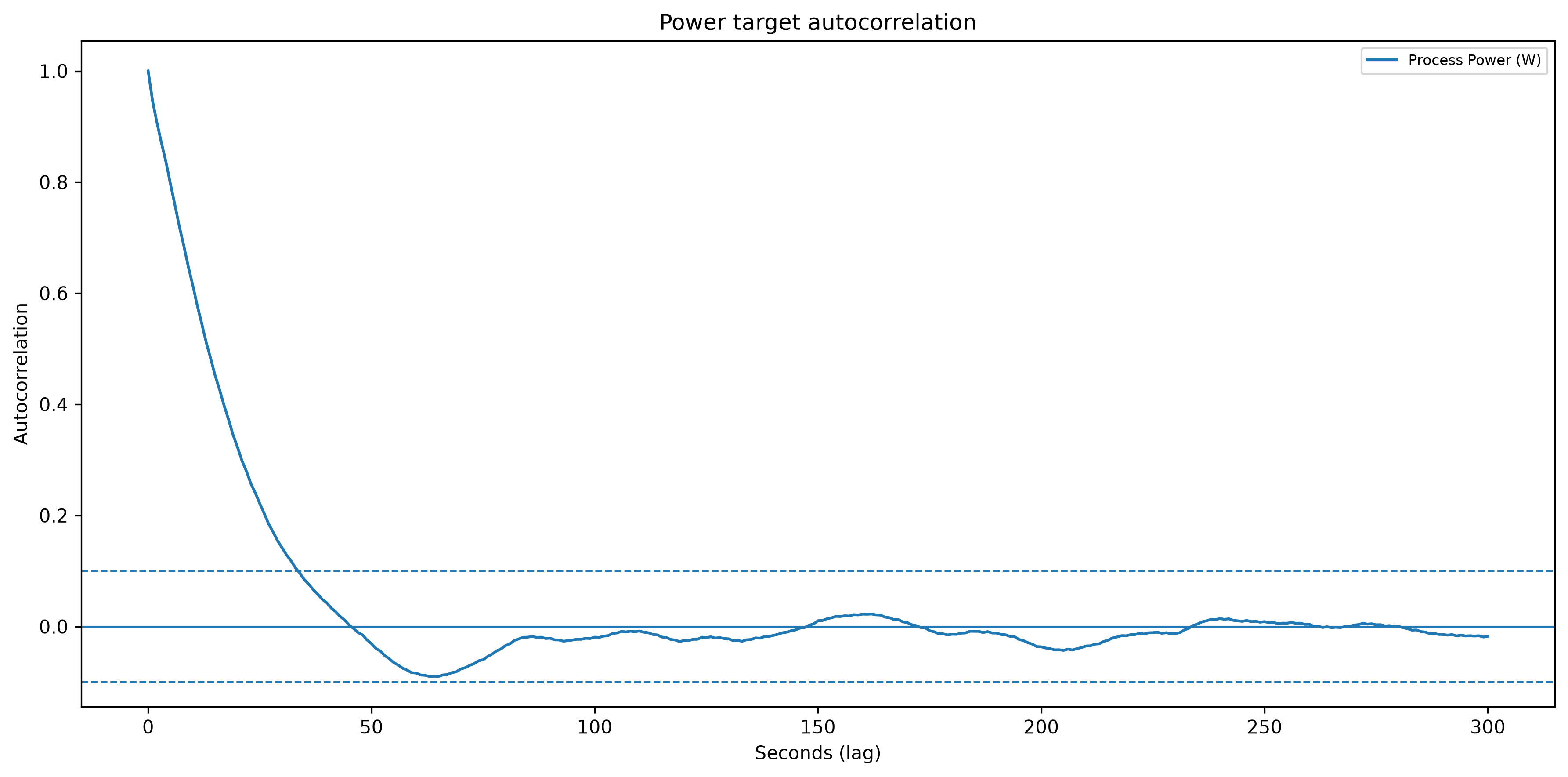}
    \label{fig:supp-acf-power}
\end{figure}

\begin{figure}[H]
    \centering
    \caption{Autocorrelation ($\rho$) of representative Glances predictors at 1~Hz sampling.
    The dashed horizontal lines indicate $|\rho|=0.1$.}
    \label{fig:supp-acf-glances}
    \includegraphics[width=0.95\textwidth]{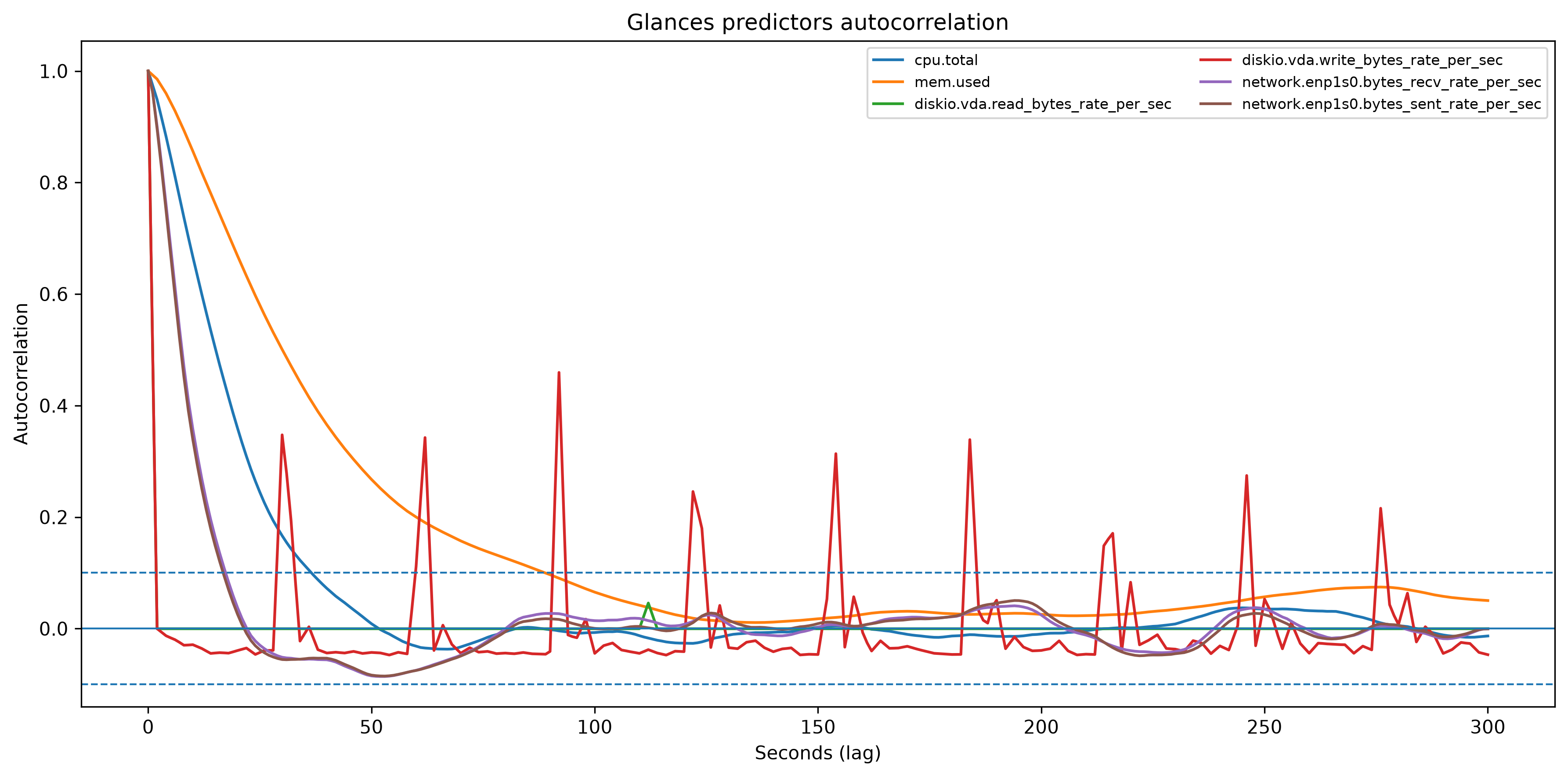}
\end{figure}

\begin{figure}[H]
    \centering
    \caption{Autocorrelation ($\rho$) of representative perf predictors at 1~Hz sampling. The dashed horizontal lines indicate $|\rho|=0.1$.}
    \includegraphics[width=0.95\textwidth]{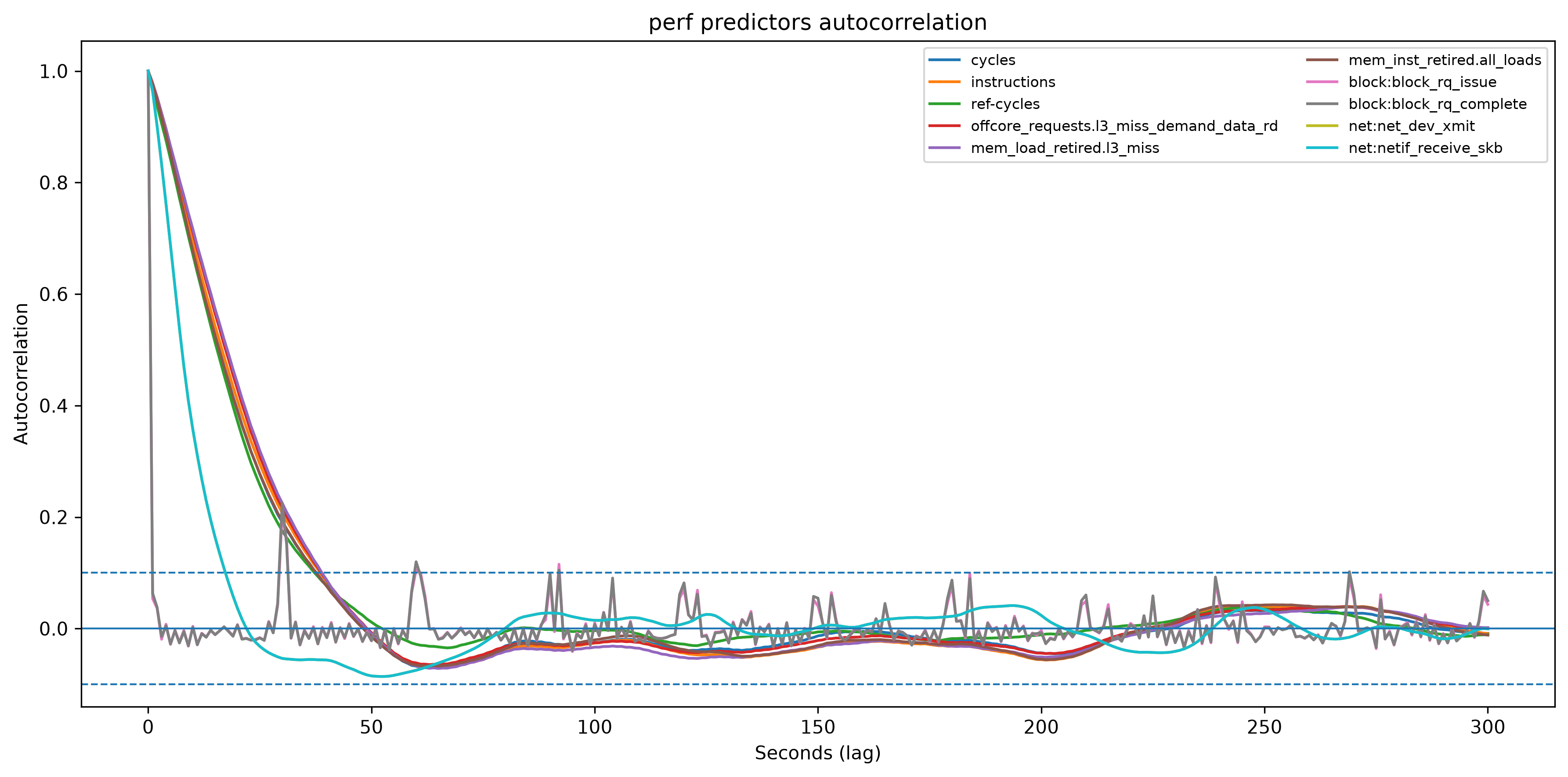}
    \label{fig:supp-acf-perf}
\end{figure}


\clearpage

\begin{figure}[H]
    \centering
    \caption{Grouped permutation importance for the Intel Xeon E3-1270 v6. Values show the decrease in held-out $R^2$ when all features in the indicated resource family are jointly permuted. The pooled column corresponds to the within-platform workload-pooled model.}
    \includegraphics[width=0.82\textwidth]{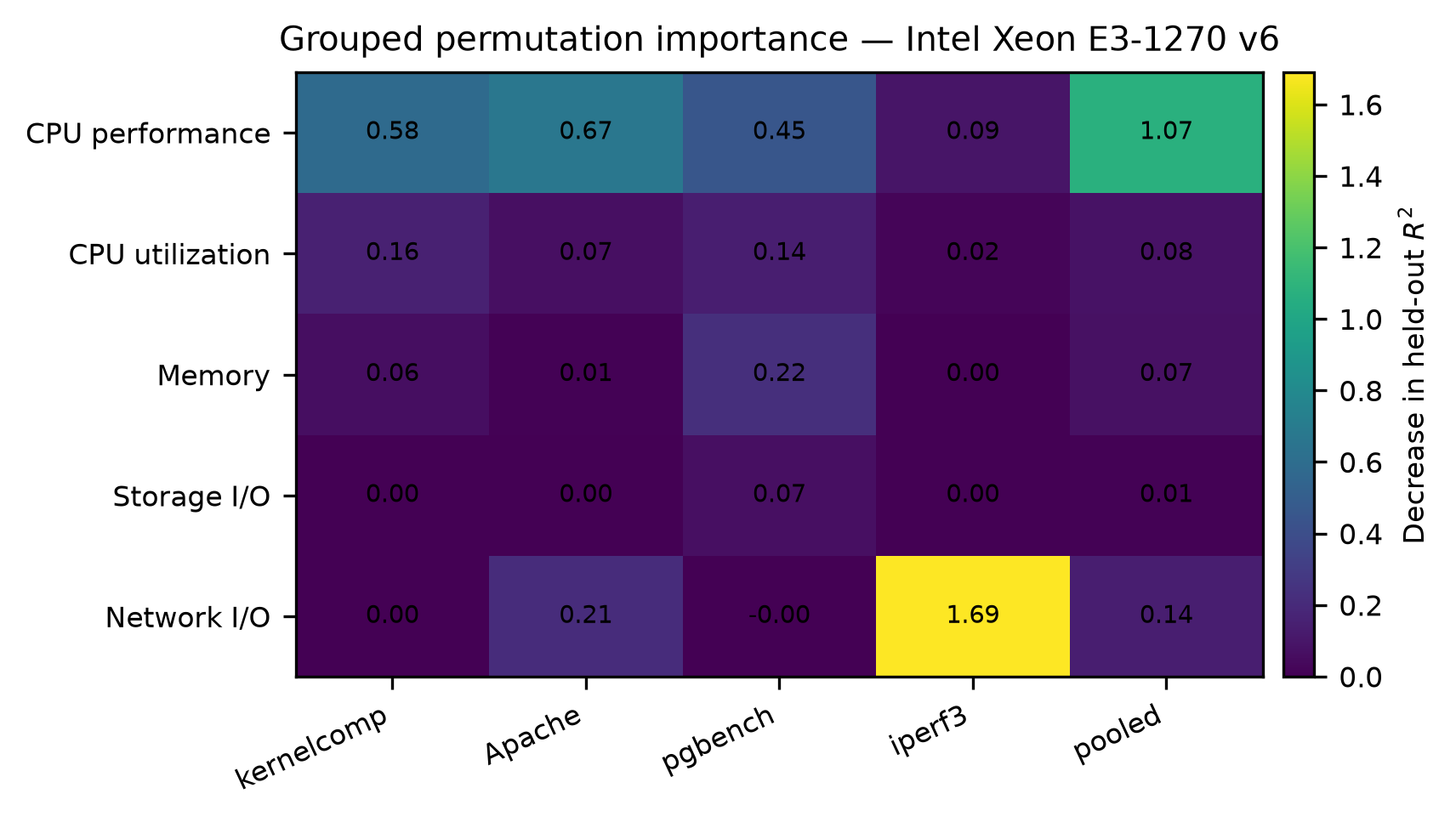}
    \label{fig:supp-grouped-intel-xeon-e3-1270-v6}
\end{figure}

\begin{figure}[H]
    \centering
    \caption{Grouped permutation importance for the Intel Xeon E-2286G. Values show the decrease in held-out $R^2$ when all features in the indicated resource family are jointly permuted. The pooled column corresponds to the within-platform workload-pooled model.}
    \includegraphics[width=0.82\textwidth]{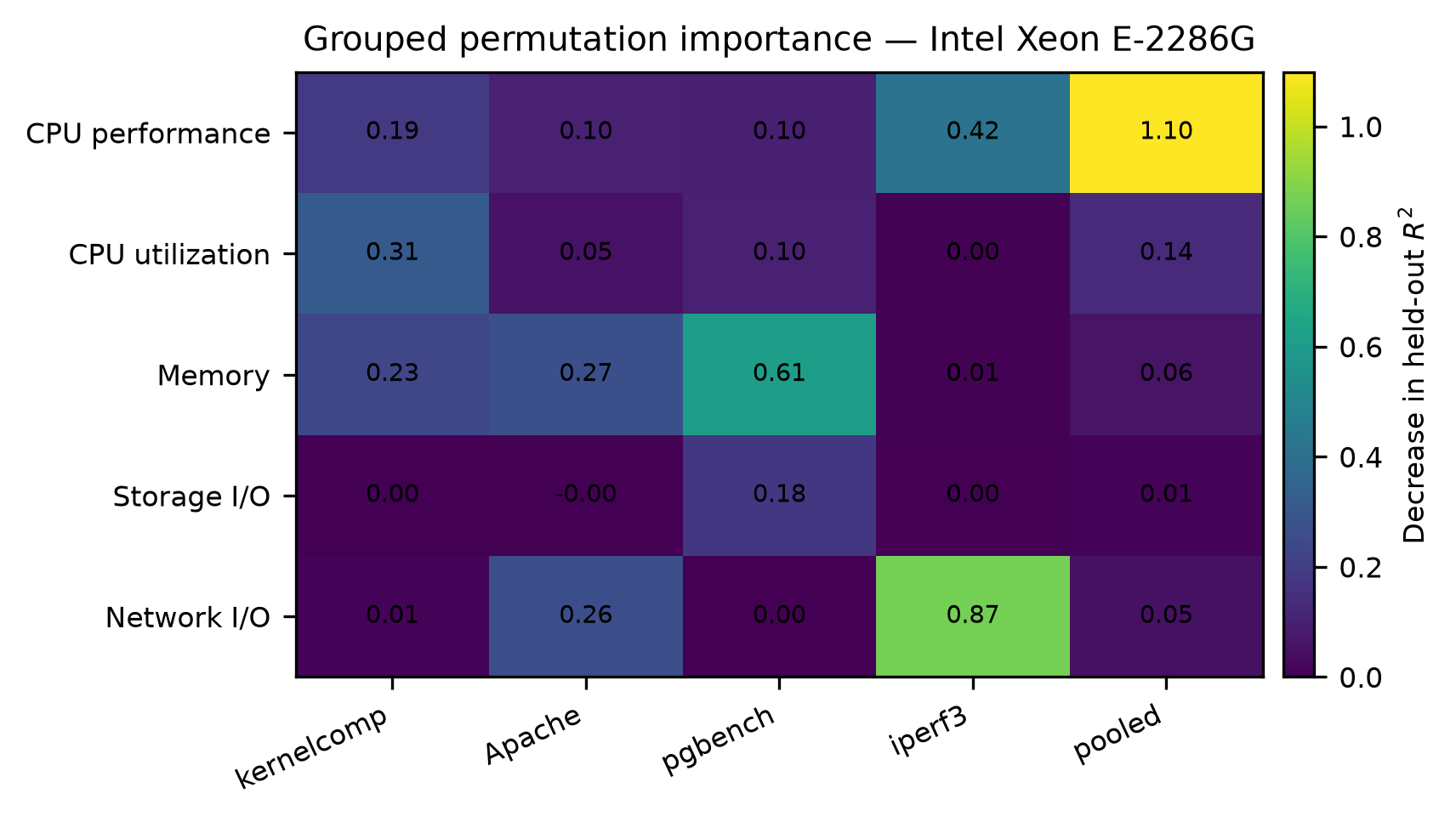}
    \label{fig:supp-grouped-intel-xeon-e-2286g}
\end{figure}

\begin{figure}[H]
    \centering
    \caption{Grouped permutation importance for the Intel Xeon 6527P. Values show the decrease in held-out $R^2$ when all features in the indicated resource family are jointly permuted. The pooled column corresponds to the within-platform workload-pooled model.}
    \includegraphics[width=0.82\textwidth]{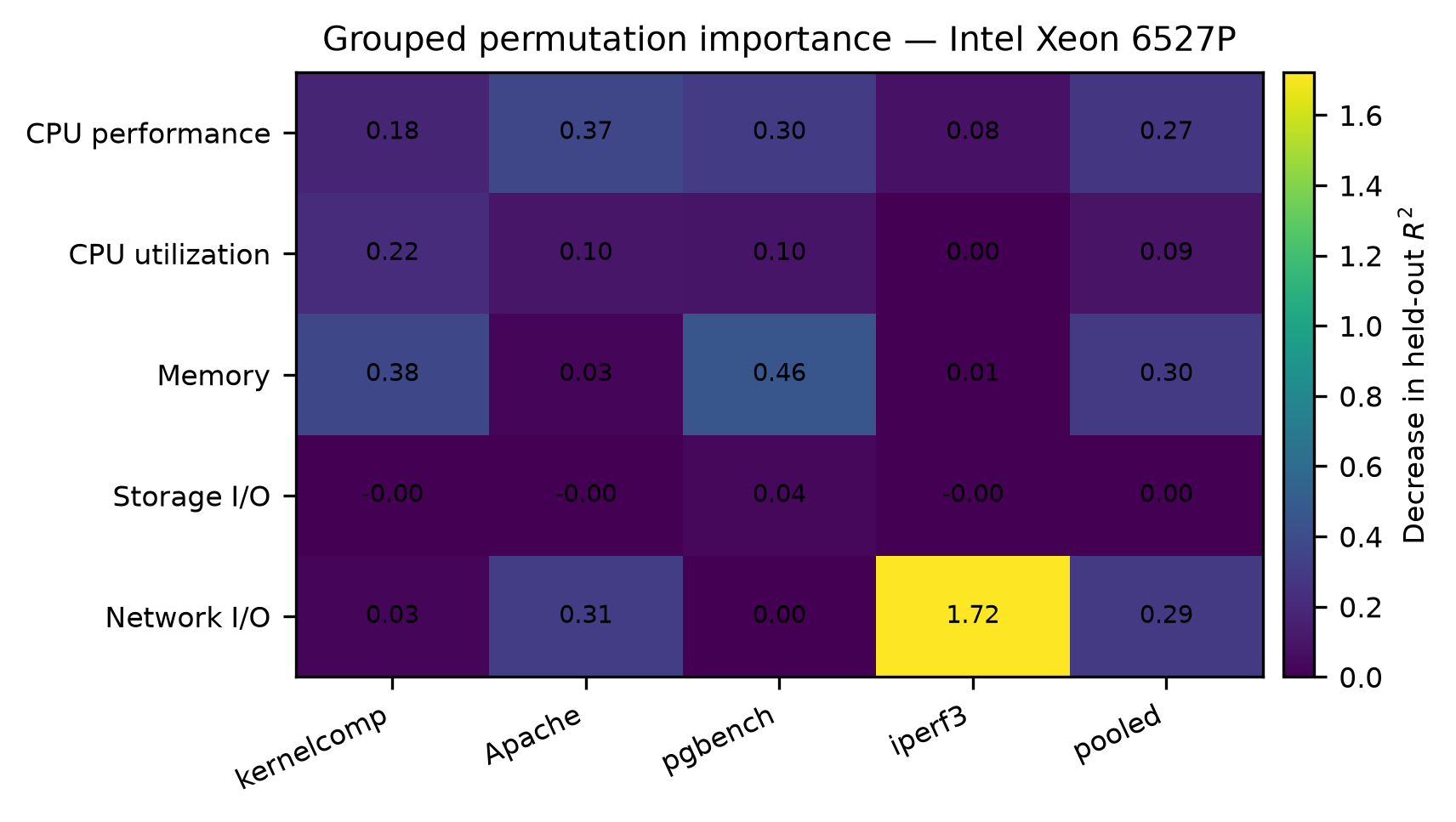}
    \label{fig:supp-grouped-intel-xeon-6527p}
\end{figure}

\begin{figure}[H]
    \centering
    \caption{Grouped permutation importance for the AMD EPYC 4245P. Values show the decrease in held-out $R^2$ when all features in the indicated resource family are jointly permuted. The pooled column corresponds to the within-platform workload-pooled model.}
    \includegraphics[width=0.82\textwidth]{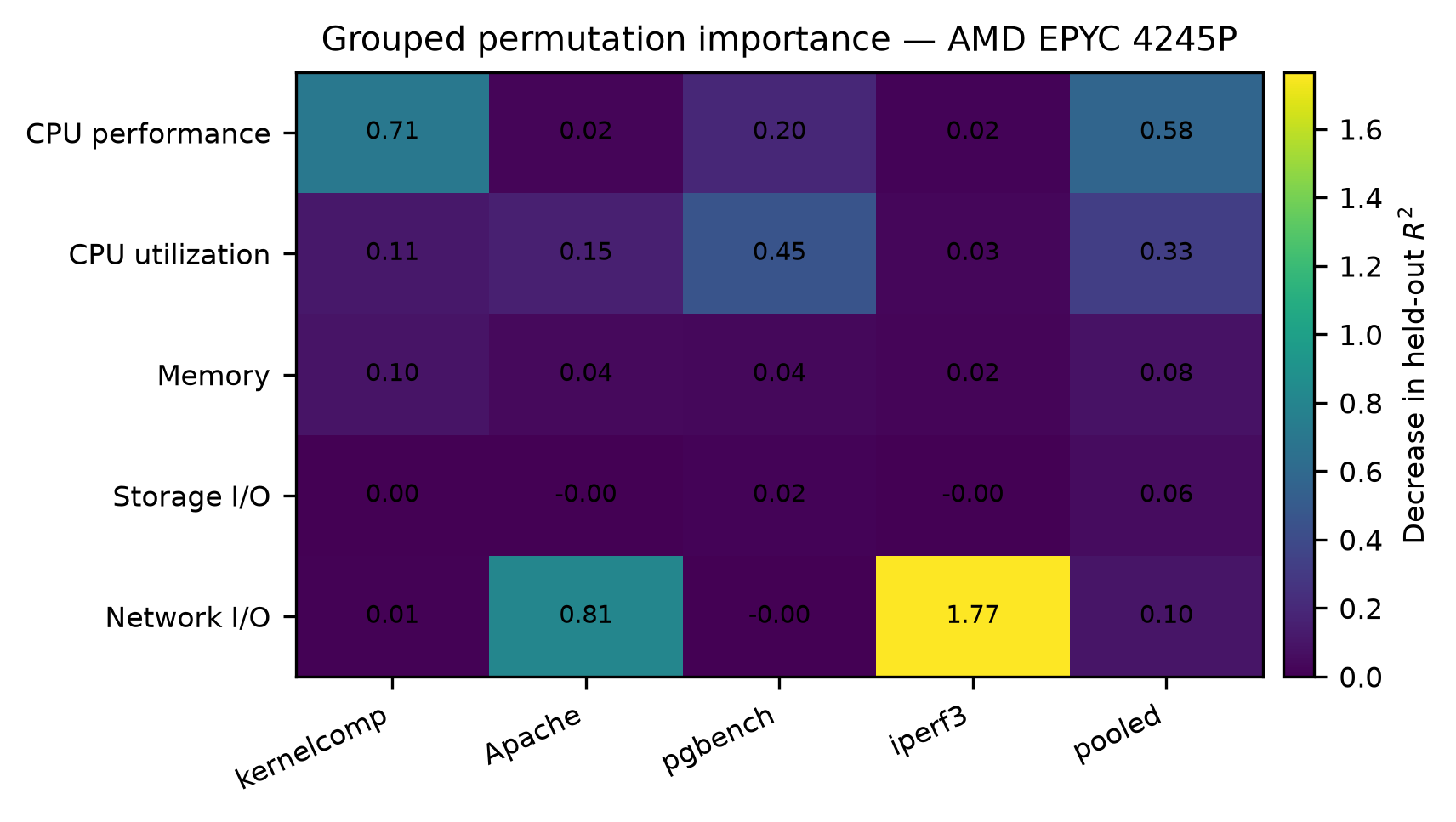}
    \label{fig:supp-grouped-amd-epyc-4245p}
\end{figure}

\begin{figure}[H]
    \centering
    \caption{Grouped permutation importance for the AMD EPYC 7443P. Values show the decrease in held-out $R^2$ when all features in the indicated resource family are jointly permuted. The pooled column corresponds to the within-platform workload-pooled model.}
    \includegraphics[width=0.82\textwidth]{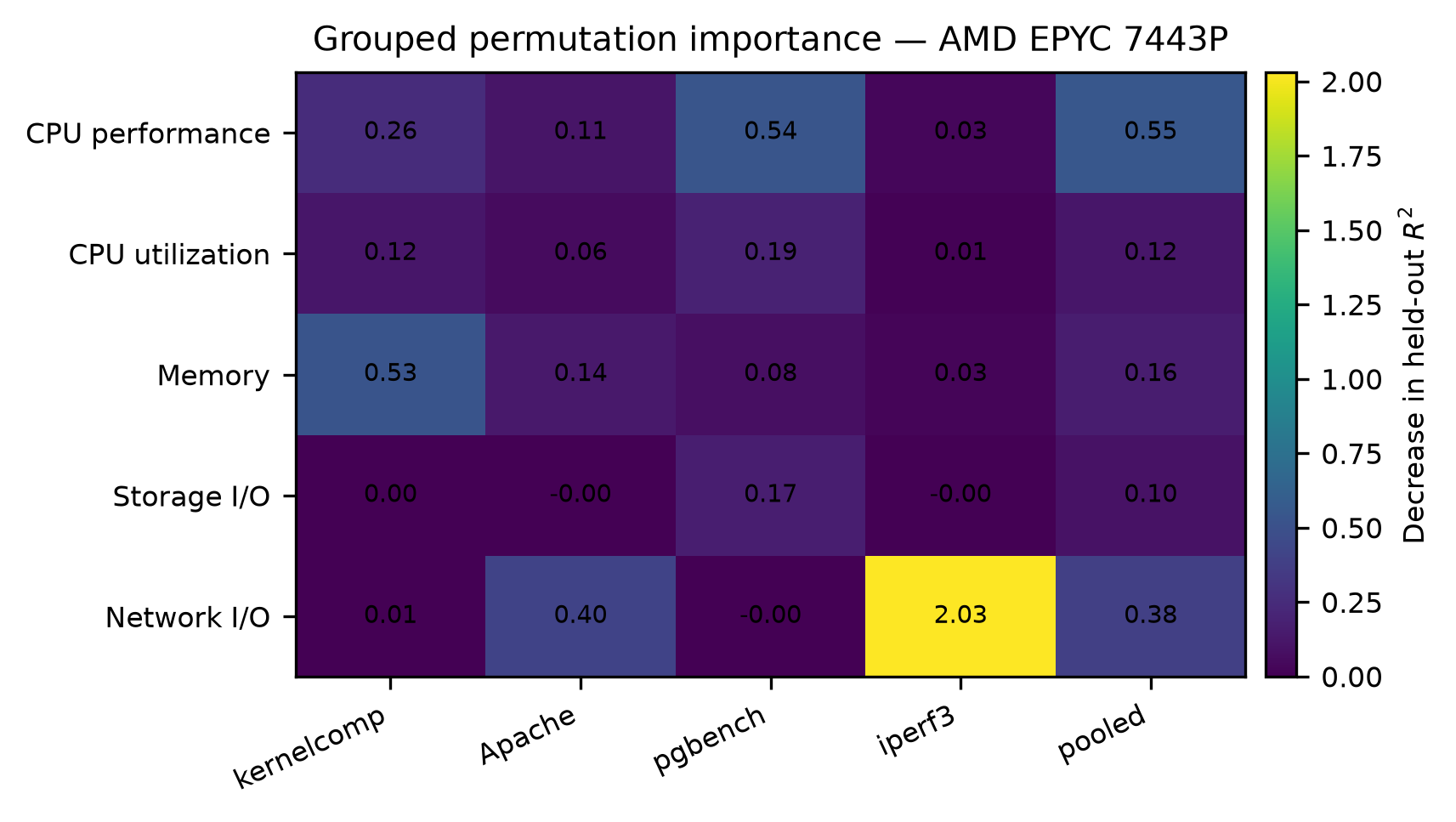}
    \label{fig:supp-grouped-amd-epyc-7443p}
\end{figure}

\begin{table}[H]
\centering
\caption{Spearman rank agreement between held-out permutation importance and GBR impurity-based feature importance.}
\label{tab:supp-importance-agreement}
\small
\setlength{\tabcolsep}{5pt}
\renewcommand{\arraystretch}{1.08}
\begin{tabular}{lccccc}
\toprule
\textbf{Platform} & \textbf{kernelcomp} & \textbf{Apache} & \textbf{pgbench} & \textbf{iperf3} & \textbf{Pooled} \\
\midrule
Intel Xeon E3-1270 v6 & 0.846 & 0.846 & 0.838 & 0.809 & 0.811 \\
Intel Xeon E-2286G & 0.939 & 0.868 & 0.681 & 0.711 & 0.762 \\
Intel Xeon 6527P & 0.892 & 0.918 & 0.922 & 0.554 & 0.907 \\
AMD EPYC 4245P & 0.888 & 0.771 & 0.862 & 0.885 & 0.874 \\
AMD EPYC 7443P & 0.964 & 0.914 & 0.889 & 0.824 & 0.836 \\
\midrule
Mean & 0.906 & 0.863 & 0.838 & 0.757 & 0.838 \\
\bottomrule
\end{tabular}
\end{table}

\begin{table}[H]
\centering
\caption{Top-five held-out permutation-important features for the Intel Xeon E3-1270 v6. Importance is the mean decrease in held-out $R^2$ after permutation of the indicated feature.}
\label{tab:supp-top5-intel-xeon-e3-1270-v6}
\footnotesize
\setlength{\tabcolsep}{4pt}
\renewcommand{\arraystretch}{1.05}
\begin{tabularx}{\textwidth}{>{\raggedright\arraybackslash}p{1.7cm} c >{\raggedright\arraybackslash}X r}
\toprule
\textbf{Workload} & \textbf{Rank} & \textbf{Feature} & \textbf{$\Delta R^2$} \\
\midrule
kernelcomp & 1 & perf.ref\_\allowbreak{}cycles\_\allowbreak{}per\_\allowbreak{}sec & 0.751 \\
 & 2 & cpu.total & 0.155 \\
 & 3 & perf.loads\_\allowbreak{}per\_\allowbreak{}sec & 0.070 \\
 & 4 & perf.cycles\_\allowbreak{}per\_\allowbreak{}sec & 0.051 \\
 & 5 & perf.instructions\_\allowbreak{}per\_\allowbreak{}sec & 0.006 \\
\addlinespace[1pt]
Apache & 1 & perf.cycles\_\allowbreak{}per\_\allowbreak{}sec & 0.754 \\
 & 2 & network.enp1s0.bytes\_\allowbreak{}recv\_\allowbreak{}rate\_\allowbreak{}per\_\allowbreak{}sec & 0.361 \\
 & 3 & cpu.total & 0.066 \\
 & 4 & perf.ref\_\allowbreak{}cycles\_\allowbreak{}per\_\allowbreak{}sec & 0.040 \\
 & 5 & perf.netif\_\allowbreak{}receive\_\allowbreak{}skb\_\allowbreak{}per\_\allowbreak{}sec & 0.032 \\
\addlinespace[1pt]
pgbench & 1 & perf.loads\_\allowbreak{}per\_\allowbreak{}sec & 0.243 \\
 & 2 & cpu.total & 0.140 \\
 & 3 & perf.cycles\_\allowbreak{}per\_\allowbreak{}sec & 0.134 \\
 & 4 & perf.ref\_\allowbreak{}cycles\_\allowbreak{}per\_\allowbreak{}sec & 0.080 \\
 & 5 & perf.block\_\allowbreak{}rq\_\allowbreak{}issue\_\allowbreak{}per\_\allowbreak{}sec & 0.039 \\
\addlinespace[1pt]
iperf3 & 1 & network.enp1s0.bytes\_\allowbreak{}recv\_\allowbreak{}rate\_\allowbreak{}per\_\allowbreak{}sec & 0.594 \\
 & 2 & perf.net\_\allowbreak{}dev\_\allowbreak{}xmit\_\allowbreak{}per\_\allowbreak{}sec & 0.472 \\
 & 3 & network.enp1s0.bytes\_\allowbreak{}sent\_\allowbreak{}rate\_\allowbreak{}per\_\allowbreak{}sec & 0.215 \\
 & 4 & perf.ref\_\allowbreak{}cycles\_\allowbreak{}per\_\allowbreak{}sec & 0.077 \\
 & 5 & perf.cycles\_\allowbreak{}per\_\allowbreak{}sec & 0.052 \\
\addlinespace[1pt]
Pooled & 1 & perf.cycles\_\allowbreak{}per\_\allowbreak{}sec & 1.610 \\
 & 2 & network.enp1s0.bytes\_\allowbreak{}sent\_\allowbreak{}rate\_\allowbreak{}per\_\allowbreak{}sec & 0.103 \\
 & 3 & perf.net\_\allowbreak{}dev\_\allowbreak{}xmit\_\allowbreak{}per\_\allowbreak{}sec & 0.095 \\
 & 4 & cpu.total & 0.084 \\
 & 5 & network.enp1s0.bytes\_\allowbreak{}recv\_\allowbreak{}rate\_\allowbreak{}per\_\allowbreak{}sec & 0.063 \\
\bottomrule
\end{tabularx}
\end{table}

\begin{table}[H]
\centering
\caption{Top-five held-out permutation-important features for the Intel Xeon E-2286G. Importance is the mean decrease in held-out $R^2$ after permutation of the indicated feature.}
\label{tab:supp-top5-intel-xeon-e-2286g}
\footnotesize
\setlength{\tabcolsep}{4pt}
\renewcommand{\arraystretch}{1.05}
\begin{tabularx}{\textwidth}{>{\raggedright\arraybackslash}p{1.7cm} c >{\raggedright\arraybackslash}X r}
\toprule
\textbf{Workload} & \textbf{Rank} & \textbf{Feature} & \textbf{$\Delta R^2$} \\
\midrule
kernelcomp & 1 & cpu.total & 0.308 \\
 & 2 & perf.ref\_\allowbreak{}cycles\_\allowbreak{}per\_\allowbreak{}sec & 0.140 \\
 & 3 & perf.dram\_\allowbreak{}read\_\allowbreak{}bytes\_\allowbreak{}est\_\allowbreak{}per\_\allowbreak{}sec & 0.066 \\
 & 4 & perf.loads\_\allowbreak{}per\_\allowbreak{}sec & 0.019 \\
 & 5 & mem.used & 0.015 \\
\addlinespace[1pt]
Apache & 1 & network.enp1s0.bytes\_\allowbreak{}recv\_\allowbreak{}rate\_\allowbreak{}per\_\allowbreak{}sec & 0.401 \\
 & 2 & perf.dram\_\allowbreak{}read\_\allowbreak{}bytes\_\allowbreak{}est\_\allowbreak{}per\_\allowbreak{}sec & 0.224 \\
 & 3 & perf.cycles\_\allowbreak{}per\_\allowbreak{}sec & 0.162 \\
 & 4 & cpu.total & 0.053 \\
 & 5 & perf.netif\_\allowbreak{}receive\_\allowbreak{}skb\_\allowbreak{}per\_\allowbreak{}sec & 0.038 \\
\addlinespace[1pt]
pgbench & 1 & perf.loads\_\allowbreak{}per\_\allowbreak{}sec & 0.305 \\
 & 2 & perf.block\_\allowbreak{}rq\_\allowbreak{}issue\_\allowbreak{}per\_\allowbreak{}sec & 0.126 \\
 & 3 & diskio.vda.write\_\allowbreak{}bytes\_\allowbreak{}rate\_\allowbreak{}per\_\allowbreak{}sec & 0.107 \\
 & 4 & cpu.total & 0.105 \\
 & 5 & perf.l3\_\allowbreak{}load\_\allowbreak{}misses\_\allowbreak{}per\_\allowbreak{}sec & 0.072 \\
\addlinespace[1pt]
iperf3 & 1 & network.enp1s0.bytes\_\allowbreak{}recv\_\allowbreak{}rate\_\allowbreak{}per\_\allowbreak{}sec & 0.802 \\
 & 2 & perf.cycles\_\allowbreak{}per\_\allowbreak{}sec & 0.565 \\
 & 3 & network.enp1s0.bytes\_\allowbreak{}sent\_\allowbreak{}rate\_\allowbreak{}per\_\allowbreak{}sec & 0.159 \\
 & 4 & perf.ref\_\allowbreak{}cycles\_\allowbreak{}per\_\allowbreak{}sec & 0.142 \\
 & 5 & perf.net\_\allowbreak{}dev\_\allowbreak{}xmit\_\allowbreak{}per\_\allowbreak{}sec & 0.073 \\
\addlinespace[1pt]
Pooled & 1 & perf.cycles\_\allowbreak{}per\_\allowbreak{}sec & 1.086 \\
 & 2 & cpu.total & 0.135 \\
 & 3 & perf.ipc & 0.093 \\
 & 4 & mem.used & 0.049 \\
 & 5 & network.enp1s0.bytes\_\allowbreak{}recv\_\allowbreak{}rate\_\allowbreak{}per\_\allowbreak{}sec & 0.029 \\
\bottomrule
\end{tabularx}
\end{table}

\begin{table}[H]
\centering
\caption{Top-five held-out permutation-important features for the Intel Xeon 6527P. Importance is the mean decrease in held-out $R^2$ after permutation of the indicated feature.}
\label{tab:supp-top5-intel-xeon-6527p}
\footnotesize
\setlength{\tabcolsep}{4pt}
\renewcommand{\arraystretch}{1.05}
\begin{tabularx}{\textwidth}{>{\raggedright\arraybackslash}p{1.7cm} c >{\raggedright\arraybackslash}X r}
\toprule 
\textbf{Workload} & \textbf{Rank} & \textbf{Feature} & \textbf{$\Delta R^2$} \\
\midrule
kernelcomp & 1 & cpu.total & 0.218 \\
 & 2 & perf.l3\_\allowbreak{}load\_\allowbreak{}misses\_\allowbreak{}per\_\allowbreak{}sec & 0.147 \\
 & 3 & perf.cycles\_\allowbreak{}per\_\allowbreak{}sec & 0.132 \\
 & 4 & mem.used & 0.048 \\
 & 5 & perf.loads\_\allowbreak{}per\_\allowbreak{}sec & 0.038 \\
\addlinespace[1pt]
Apache & 1 & perf.cycles\_\allowbreak{}per\_\allowbreak{}sec & 0.354 \\
 & 2 & network.enp1s0.bytes\_\allowbreak{}sent\_\allowbreak{}rate\_\allowbreak{}per\_\allowbreak{}sec & 0.252 \\
 & 3 & network.enp1s0.bytes\_\allowbreak{}recv\_\allowbreak{}rate\_\allowbreak{}per\_\allowbreak{}sec & 0.124 \\
 & 4 & cpu.total & 0.104 \\
 & 5 & perf.loads\_\allowbreak{}per\_\allowbreak{}sec & 0.012 \\
\addlinespace[1pt]
pgbench & 1 & perf.instructions\_\allowbreak{}per\_\allowbreak{}sec & 0.267 \\
 & 2 & perf.dram\_\allowbreak{}read\_\allowbreak{}bytes\_\allowbreak{}est\_\allowbreak{}per\_\allowbreak{}sec & 0.162 \\
 & 3 & cpu.total & 0.098 \\
 & 4 & perf.l3\_\allowbreak{}load\_\allowbreak{}misses\_\allowbreak{}per\_\allowbreak{}sec & 0.063 \\
 & 5 & perf.loads\_\allowbreak{}per\_\allowbreak{}sec & 0.027 \\
\addlinespace[1pt]
iperf3 & 1 & network.enp1s0.bytes\_\allowbreak{}recv\_\allowbreak{}rate\_\allowbreak{}per\_\allowbreak{}sec & 0.541 \\
 & 2 & perf.netif\_\allowbreak{}receive\_\allowbreak{}skb\_\allowbreak{}per\_\allowbreak{}sec & 0.228 \\
 & 3 & perf.cycles\_\allowbreak{}per\_\allowbreak{}sec & 0.107 \\
 & 4 & perf.net\_\allowbreak{}dev\_\allowbreak{}xmit\_\allowbreak{}per\_\allowbreak{}sec & 0.097 \\
 & 5 & perf.ref\_\allowbreak{}cycles\_\allowbreak{}per\_\allowbreak{}sec & 0.031 \\
\addlinespace[1pt]
Pooled & 1 & perf.cycles\_\allowbreak{}per\_\allowbreak{}sec & 0.143 \\
 & 2 & network.enp1s0.bytes\_\allowbreak{}recv\_\allowbreak{}rate\_\allowbreak{}per\_\allowbreak{}sec & 0.138 \\
 & 3 & perf.l3\_\allowbreak{}load\_\allowbreak{}misses\_\allowbreak{}per\_\allowbreak{}sec & 0.114 \\
 & 4 & cpu.total & 0.089 \\
 & 5 & perf.ref\_\allowbreak{}cycles\_\allowbreak{}per\_\allowbreak{}sec & 0.068 \\
\bottomrule
\end{tabularx}
\end{table}

\begin{table}[H]
\centering
\caption{Top-five held-out permutation-important features for the AMD EPYC 4245P. Importance is the mean decrease in held-out $R^2$ after permutation of the indicated feature.}
\label{tab:supp-top5-amd-epyc-4245p}
\footnotesize
\setlength{\tabcolsep}{4pt}
\renewcommand{\arraystretch}{1.05}
\begin{tabularx}{\textwidth}{>{\raggedright\arraybackslash}p{1.7cm} c >{\raggedright\arraybackslash}X r}
\toprule
\textbf{Workload} & \textbf{Rank} & \textbf{Feature} & \textbf{$\Delta R^2$} \\
\midrule
kernelcomp & 1 & perf.cycles\_\allowbreak{}per\_\allowbreak{}sec & 0.178 \\
 & 2 & perf.ref\_\allowbreak{}cycles\_\allowbreak{}per\_\allowbreak{}sec & 0.140 \\
 & 3 & cpu.total & 0.112 \\
 & 4 & perf.loads\_\allowbreak{}dispatched\_\allowbreak{}per\_\allowbreak{}sec & 0.079 \\
 & 5 & perf.instructions\_\allowbreak{}per\_\allowbreak{}sec & 0.029 \\
\addlinespace[1pt]
Apache & 1 & network.enp1s0.bytes\_\allowbreak{}recv\_\allowbreak{}rate\_\allowbreak{}per\_\allowbreak{}sec & 0.691 \\
 & 2 & cpu.total & 0.155 \\
 & 3 & network.enp1s0.bytes\_\allowbreak{}sent\_\allowbreak{}rate\_\allowbreak{}per\_\allowbreak{}sec & 0.028 \\
 & 4 & perf.loads\_\allowbreak{}dispatched\_\allowbreak{}per\_\allowbreak{}sec & 0.021 \\
 & 5 & perf.ref\_\allowbreak{}cycles\_\allowbreak{}per\_\allowbreak{}sec & 0.013 \\
\addlinespace[1pt]
pgbench & 1 & cpu.total & 0.449 \\
 & 2 & perf.instructions\_\allowbreak{}per\_\allowbreak{}sec & 0.033 \\
 & 3 & perf.sys\_\allowbreak{}dram\_\allowbreak{}io\_\allowbreak{}fills\_\allowbreak{}per\_\allowbreak{}sec & 0.029 \\
 & 4 & perf.ref\_\allowbreak{}cycles\_\allowbreak{}per\_\allowbreak{}sec & 0.028 \\
 & 5 & perf.ipc & 0.017 \\
\addlinespace[1pt]
iperf3 & 1 & network.enp1s0.bytes\_\allowbreak{}recv\_\allowbreak{}rate\_\allowbreak{}per\_\allowbreak{}sec & 0.660 \\
 & 2 & perf.netif\_\allowbreak{}receive\_\allowbreak{}skb\_\allowbreak{}per\_\allowbreak{}sec & 0.085 \\
 & 3 & network.enp1s0.bytes\_\allowbreak{}sent\_\allowbreak{}rate\_\allowbreak{}per\_\allowbreak{}sec & 0.048 \\
 & 4 & perf.net\_\allowbreak{}dev\_\allowbreak{}xmit\_\allowbreak{}per\_\allowbreak{}sec & 0.037 \\
 & 5 & perf.instructions\_\allowbreak{}per\_\allowbreak{}sec & 0.033 \\
\addlinespace[1pt]
Pooled & 1 & cpu.total & 0.327 \\
 & 2 & perf.ref\_\allowbreak{}cycles\_\allowbreak{}per\_\allowbreak{}sec & 0.279 \\
 & 3 & perf.cycles\_\allowbreak{}per\_\allowbreak{}sec & 0.083 \\
 & 4 & network.enp1s0.bytes\_\allowbreak{}recv\_\allowbreak{}rate\_\allowbreak{}per\_\allowbreak{}sec & 0.054 \\
 & 5 & perf.sys\_\allowbreak{}dram\_\allowbreak{}io\_\allowbreak{}fills\_\allowbreak{}per\_\allowbreak{}sec & 0.042 \\
\bottomrule
\end{tabularx}
\end{table}

\begin{table}[H]
\centering
\caption{Top-five held-out permutation-important features for the AMD EPYC 7443P. Importance is the mean decrease in held-out $R^2$ after permutation of the indicated feature.}
\label{tab:supp-top5-amd-epyc-7443p}
\footnotesize
\setlength{\tabcolsep}{4pt}
\renewcommand{\arraystretch}{1.05}
\begin{tabularx}{\textwidth}{>{\raggedright\arraybackslash}p{1.7cm} c >{\raggedright\arraybackslash}X r}
\toprule
\textbf{Workload} & \textbf{Rank} & \textbf{Feature} & \textbf{$\Delta R^2$} \\
\midrule
kernelcomp & 1 & perf.loads\_\allowbreak{}dispatched\_\allowbreak{}per\_\allowbreak{}sec & 0.302 \\
 & 2 & perf.cycles\_\allowbreak{}per\_\allowbreak{}sec & 0.282 \\
 & 3 & cpu.total & 0.121 \\
 & 4 & perf.sys\_\allowbreak{}dram\_\allowbreak{}io\_\allowbreak{}fills\_\allowbreak{}per\_\allowbreak{}sec & 0.086 \\
 & 5 & mem.used & 0.015 \\
\addlinespace[1pt]
Apache & 1 & network.enp1s0.bytes\_\allowbreak{}recv\_\allowbreak{}rate\_\allowbreak{}per\_\allowbreak{}sec & 0.359 \\
 & 2 & network.enp1s0.bytes\_\allowbreak{}sent\_\allowbreak{}rate\_\allowbreak{}per\_\allowbreak{}sec & 0.139 \\
 & 3 & perf.cycles\_\allowbreak{}per\_\allowbreak{}sec & 0.121 \\
 & 4 & perf.loads\_\allowbreak{}dispatched\_\allowbreak{}per\_\allowbreak{}sec & 0.100 \\
 & 5 & cpu.total & 0.060 \\
\addlinespace[1pt]
pgbench & 1 & perf.cycles\_\allowbreak{}per\_\allowbreak{}sec & 0.335 \\
 & 2 & cpu.total & 0.185 \\
 & 3 & perf.loads\_\allowbreak{}dispatched\_\allowbreak{}per\_\allowbreak{}sec & 0.055 \\
 & 4 & perf.ipc & 0.051 \\
 & 5 & diskio.vda.write\_\allowbreak{}bytes\_\allowbreak{}rate\_\allowbreak{}per\_\allowbreak{}sec & 0.050 \\
\addlinespace[1pt]
iperf3 & 1 & network.enp1s0.bytes\_\allowbreak{}recv\_\allowbreak{}rate\_\allowbreak{}per\_\allowbreak{}sec & 0.721 \\
 & 2 & network.enp1s0.bytes\_\allowbreak{}sent\_\allowbreak{}rate\_\allowbreak{}per\_\allowbreak{}sec & 0.156 \\
 & 3 & perf.net\_\allowbreak{}dev\_\allowbreak{}xmit\_\allowbreak{}per\_\allowbreak{}sec & 0.102 \\
 & 4 & perf.netif\_\allowbreak{}receive\_\allowbreak{}skb\_\allowbreak{}per\_\allowbreak{}sec & 0.076 \\
 & 5 & perf.cycles\_\allowbreak{}per\_\allowbreak{}sec & 0.019 \\
\addlinespace[1pt]
Pooled & 1 & perf.cycles\_\allowbreak{}per\_\allowbreak{}sec & 0.476 \\
 & 2 & cpu.total & 0.122 \\
 & 3 & network.enp1s0.bytes\_\allowbreak{}recv\_\allowbreak{}rate\_\allowbreak{}per\_\allowbreak{}sec & 0.102 \\
 & 4 & perf.sys\_\allowbreak{}dram\_\allowbreak{}io\_\allowbreak{}fills\_\allowbreak{}per\_\allowbreak{}sec & 0.084 \\
 & 5 & perf.net\_\allowbreak{}dev\_\allowbreak{}xmit\_\allowbreak{}per\_\allowbreak{}sec & 0.075 \\
\bottomrule
\end{tabularx}
\end{table}

\FloatBarrier

\begin{table}[tb]
\centering
\caption{Per-workload leave-one-hardware-out (LOHO) prediction
performance. Each platform is excluded entirely from training,
and the model is evaluated separately on its four workloads.}
\label{tab:supp-loho-workloads}
\small
\setlength{\tabcolsep}{4pt}
\renewcommand{\arraystretch}{1.1}

\begin{tabular}{llrrrrrr}
\toprule
\textbf{Held-out platform} &
\textbf{Workload} & 
$\mathbf{R^2}$ &
\textbf{MAE} &
\textbf{RMSE} &
\textbf{NMAE} &
\textbf{Bias} &
$\mathbf{n_{\mathrm{test}}}$ \\
&
&
&
\textbf{(W)} &
\textbf{(W)} &
\textbf{(\%)} &
\textbf{(W)} & \\
\midrule

E3-1270 v6 & kernelcomp &  0.838 &  5.394 &  5.749 & 17.58 &   2.238 & 10947 \\
           & Apache     &  0.822 &  5.430 &  6.712 & 21.29 &  -0.746 & 10933 \\
           & pgbench    & -1.431 & 14.128 & 15.974 & 46.67 &  13.754 & 11094 \\
           & iperf3     &  0.293 &  5.499 &  8.528 & 53.25 &   3.065 & 11387 \\
\midrule

E-2286G    & kernelcomp &  0.385 & 14.174 & 16.566 & 32.72 & -13.936 & 10825 \\
           & Apache     &  0.255 & 16.039 & 20.137 & 43.98 & -14.085 & 10708 \\
           & pgbench    & -1.169 & 21.775 & 23.633 & 41.23 & -21.388 & 10575 \\
           & iperf3     &  0.685 &  6.149 &  9.004 & 56.99 &  -0.411 & 10634 \\
\midrule

Xeon 6527P & kernelcomp &  0.549 &  5.358 &  7.382 & 23.83 &   2.432 & 11750 \\
           & Apache     & -2.399 &  9.370 & 12.951 & 53.50 &   8.056 & 10922 \\
           & pgbench    & -2.890 & 10.008 & 13.902 & 46.47 &   9.531 & 10636 \\
           & iperf3     &  0.298 &  6.466 &  8.266 & 38.93 &  -5.742 & 11042 \\
\midrule

EPYC 4245P & kernelcomp &  0.927 &  3.771 &  5.014 & 12.04 &   2.895 & 11945 \\
           & Apache     &  0.560 &  9.020 & 10.599 & 24.52 &  -8.119 & 10982 \\
           & pgbench    &  0.626 & 10.192 & 12.781 & 23.61 &  -9.757 & 12851 \\
           & iperf3     &  0.378 &  5.970 &  7.032 & 38.98 &  -3.863 & 12106 \\
\midrule

EPYC 7443P & kernelcomp &  0.716 &  5.852 &  6.503 & 26.83 &   5.800 & 11151 \\
           & Apache     & -0.383 &  7.928 &  9.471 & 38.03 &   7.496 & 10582 \\
           & pgbench    &  0.174 &  6.424 &  7.410 & 20.00 &   5.550 & 10591 \\
           & iperf3     &  0.838 &  3.061 &  4.322 & 19.40 &   0.538 & 11474 \\

\bottomrule
\end{tabular}
\end{table}

\begin{table}[htbp]
\centering
\caption{Within-platform feature-set (F1-F5) ablation for selected Intel Xeon
hardware--workload combinations. Values report mean $R^2$ across
100 temporally blocked train--test splits with 120-s exclusion gaps.}
\label{tab:supp-feature-ablation}

\small
\setlength{\tabcolsep}{6pt}
\renewcommand{\arraystretch}{1.1}

\begin{tabular}{llccccc}
\toprule
\textbf{Platform} &
\textbf{Workload} &
\textbf{F1 $\mathbf{R^2}$} &
\textbf{F2 $\mathbf{R^2}$} &
\textbf{F3 $\mathbf{R^2}$} &
\textbf{F4 $\mathbf{R^2}$} &
\textbf{F5 $\mathbf{R^2}$} \\
\midrule
E3-1270 v6 & kernelcomp & 0.960 & 0.984 & 0.984 & 0.962 & 0.985 \\
           & Apache     & 0.877 & 0.932 & 0.927 & 0.879 & 0.936 \\
           & pgbench    & 0.649 & 0.861 & 0.861 & 0.806 & 0.857 \\
\midrule
E-2286G    & kernelcomp & 0.927 & 0.959 & 0.959 & 0.934 & 0.961 \\
           & Apache     & 0.899 & 0.942 & 0.935 & 0.901 & 0.945 \\
           & pgbench    & 0.693 & 0.789 & 0.787 & 0.738 & 0.786 \\
           & iperf3     & 0.879 & 0.906 & 0.891 & 0.901 & 0.916 \\
\bottomrule
\end{tabular}
\end{table}

\begin{table}[t]
\centering
\caption{Model-inference power overhead on the Intel Xeon
E3-1270 v6 under kernel compilation. Each matched pair
consists of two 300-s runs with identical configurations,
differing only in whether model inference is enabled
(ON) or disabled (OFF). Guest-side metric collection
remains enabled in both conditions. Power denotes
JoularCore VM-attributed process power.}
\label{tab:inference-overhead}
\small
\begin{tabular}{lrrrr}
\hline
\textbf{Pair}
& \textbf{OFF (W)}
& \textbf{ON (W)}
& $\boldsymbol{\Delta P}$ \textbf{(W)}
& $\boldsymbol{\Delta P}$ \textbf{(\%)} \\
\hline
R01 & 42.793 & 42.965 & +0.173 & +0.404 \\
R02 & 42.852 & 42.920 & +0.068 & +0.158 \\
R03 & 42.765 & 42.820 & +0.056 & +0.130 \\
\hline
Mean & 42.803 & 42.902 & +0.099 & +0.231 \\
\hline
\multicolumn{3}{l}{95\% CI (paired difference)}
& $[-0.061,\,0.259]$
& $[-0.143,\,0.604]$ \\
\hline
\end{tabular}
\end{table}
 
\begin{table}[t]
\centering
\caption{Inference performance across the three 300-s
inference-enabled measurement windows ($n=897$
predictions). Latencies are descriptive summaries of
individual predictions. Observation age denotes
the elapsed time between the source telemetry
timestamp and completion of the corresponding
power prediction.}
\label{tab:inference-latency}
\small
\begin{tabular}{lrrr}
\hline 
\textbf{Metric} 
& \textbf{Mean}
& \textbf{Median}
& \textbf{P95} \\
\hline
F5 feature assembly (ms) & 0.451 & 0.439 & 0.473 \\
GBR prediction (ms) & 1.370 & 1.247 & 2.351 \\
Observation age (ms) & 1544 & 1544 & 1732 \\
\hline
\end{tabular}
\end{table}

\clearpage

\begin{figure}[H]
    \centering
    \caption{SHAP summary for the within-platform workload-pooled model on the Intel Xeon E3-1270 v6. Each point represents one held-out observation; horizontal position gives the SHAP contribution to predicted VM-associated power, and color denotes the corresponding feature value.}
    \includegraphics[width=0.82\textwidth]{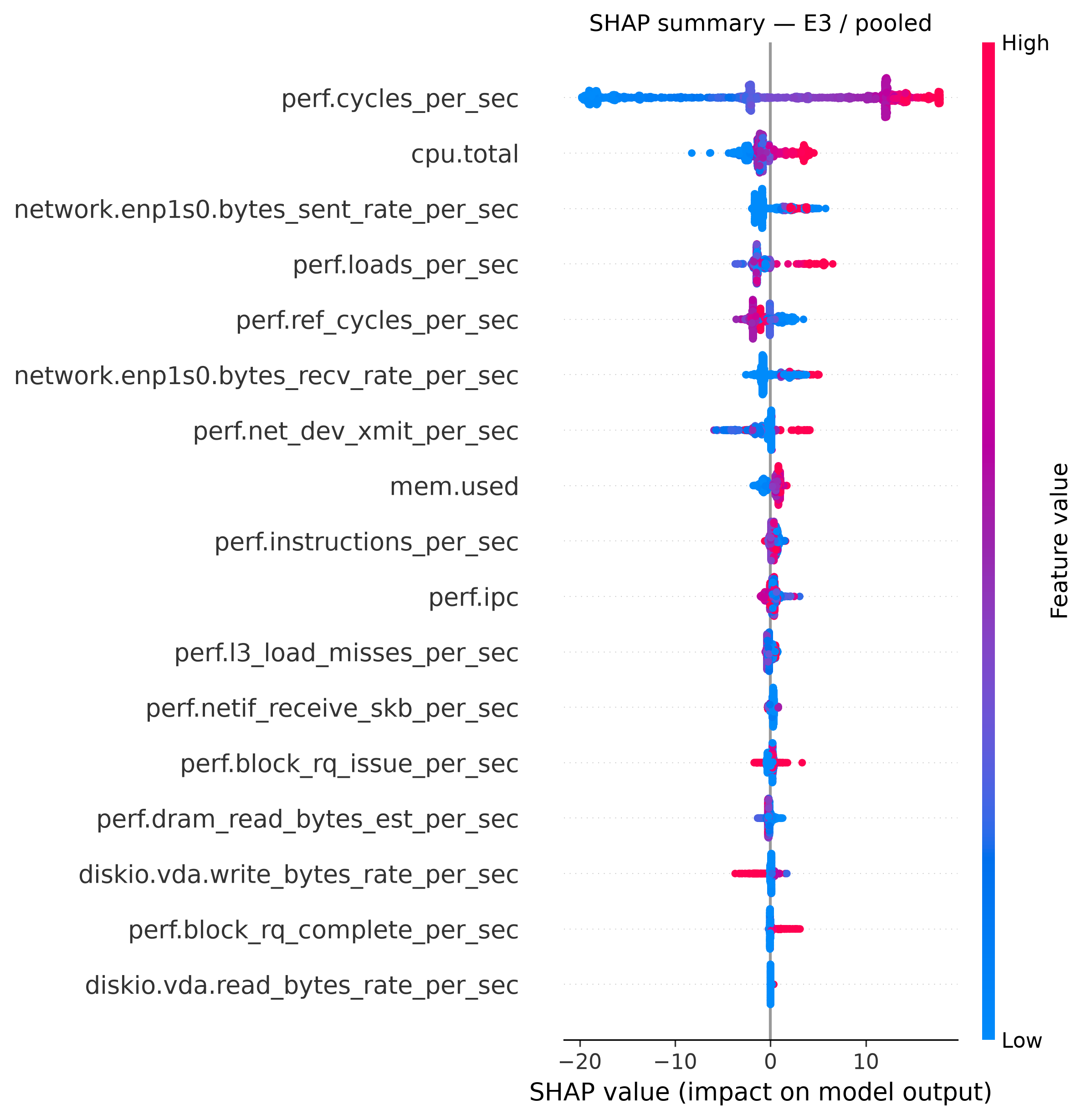}
    \label{fig:supp-shap-pooled-intel-xeon-e3-1270-v6}
\end{figure}

\begin{figure}[H]
    \centering
    \caption{SHAP summary for the within-platform workload-pooled model on the Intel Xeon E-2286G. Each point represents one held-out observation; horizontal position gives the SHAP contribution to predicted VM-associated power, and color denotes the corresponding feature value.}
    \includegraphics[width=0.82\textwidth]{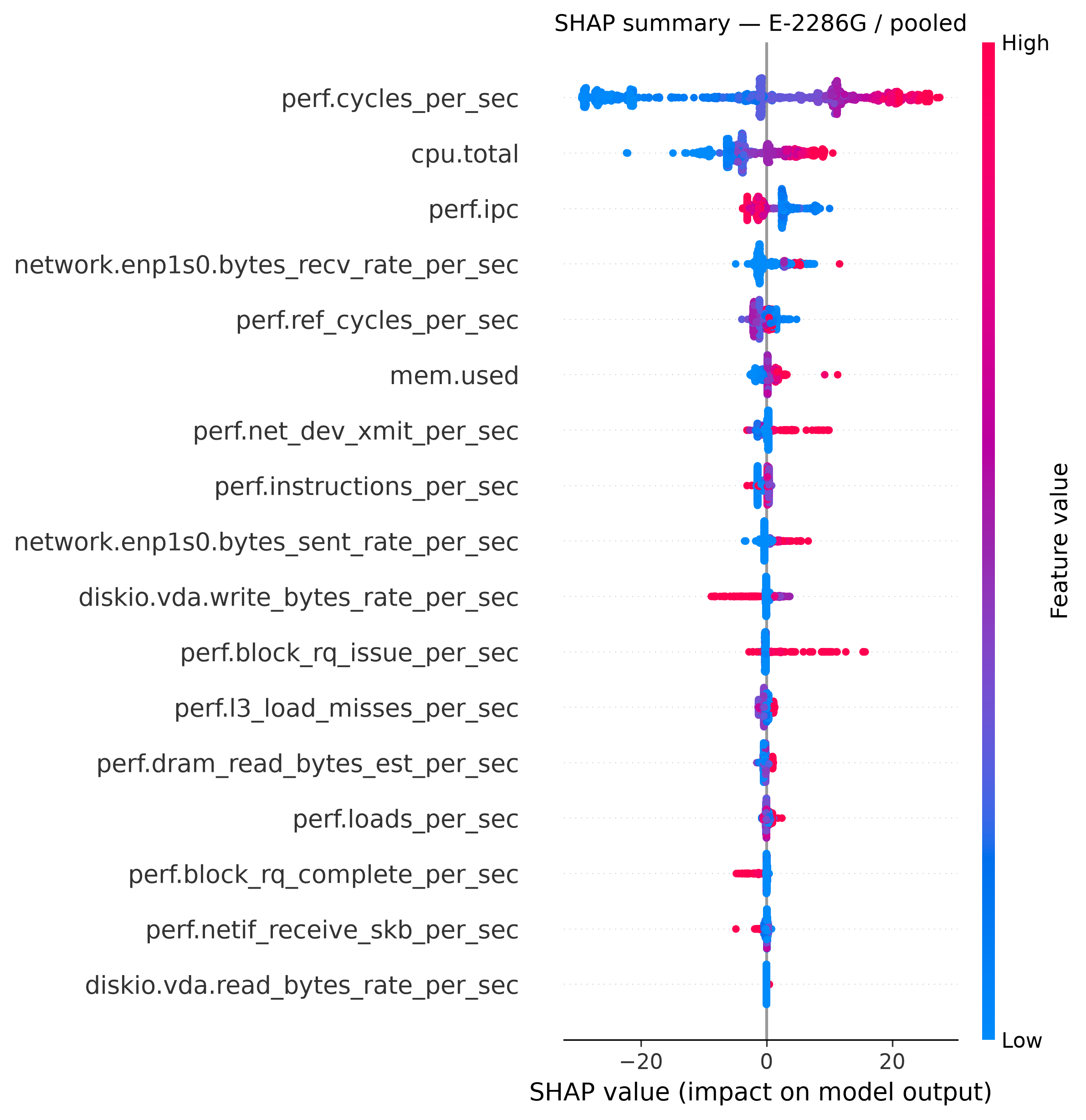}
    \label{fig:supp-shap-pooled-intel-xeon-e-2286g}
\end{figure}

\begin{figure}[H]
    \centering
    \caption{SHAP summary for the within-platform workload-pooled model on the Intel Xeon 6527P. Each point represents one held-out observation; horizontal position gives the SHAP contribution to predicted VM-associated power, and color denotes the corresponding feature value.}
    \includegraphics[width=0.82\textwidth]{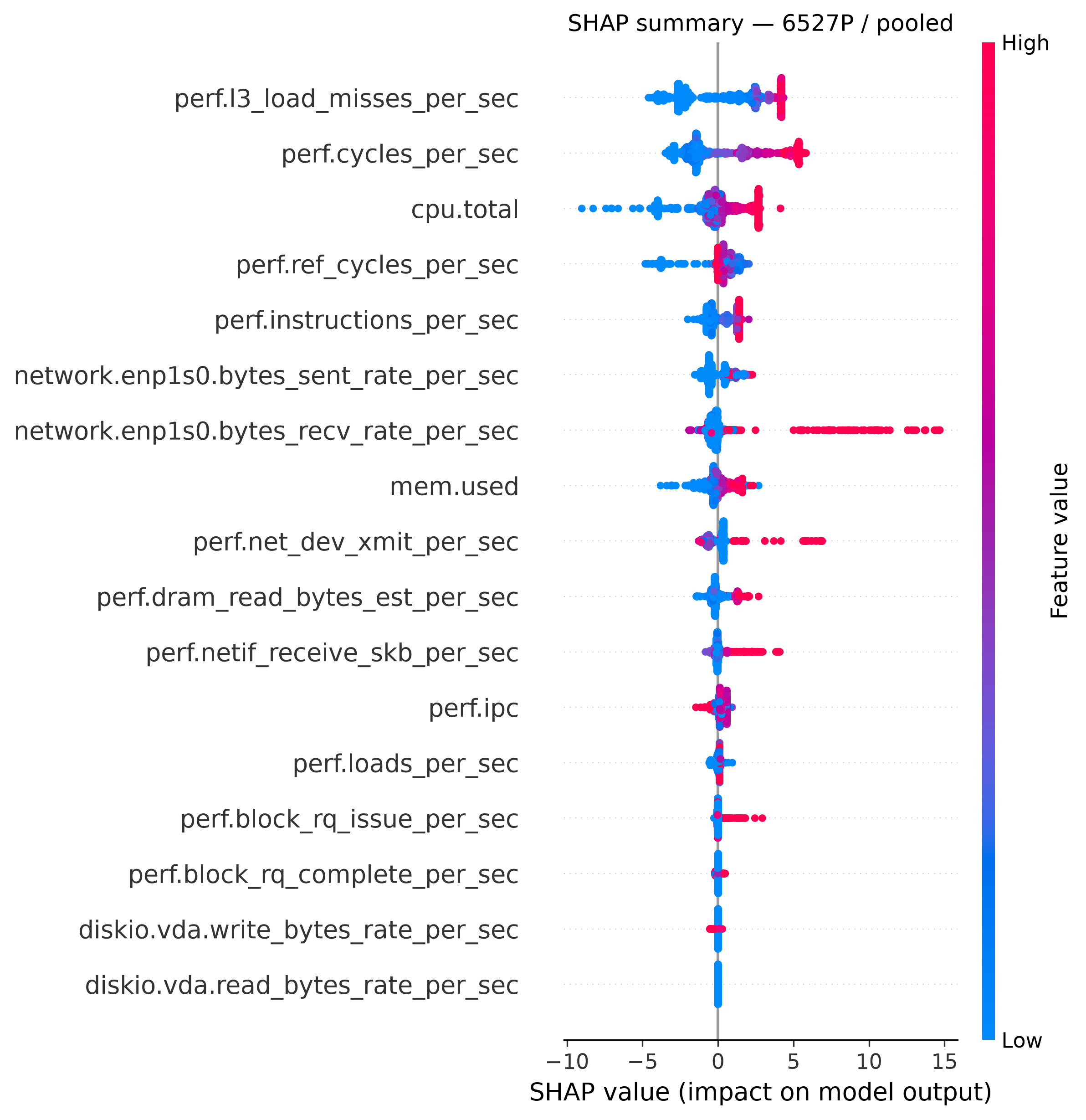}
    \label{fig:supp-shap-pooled-intel-xeon-6527p}
\end{figure}

\begin{figure}[H]
    \centering
    \caption{SHAP summary for the within-platform workload-pooled model on the AMD EPYC 4245P. Each point represents one held-out observation; horizontal position gives the SHAP contribution to predicted VM-associated power, and color denotes the corresponding feature value.}
    \includegraphics[width=0.82\textwidth]{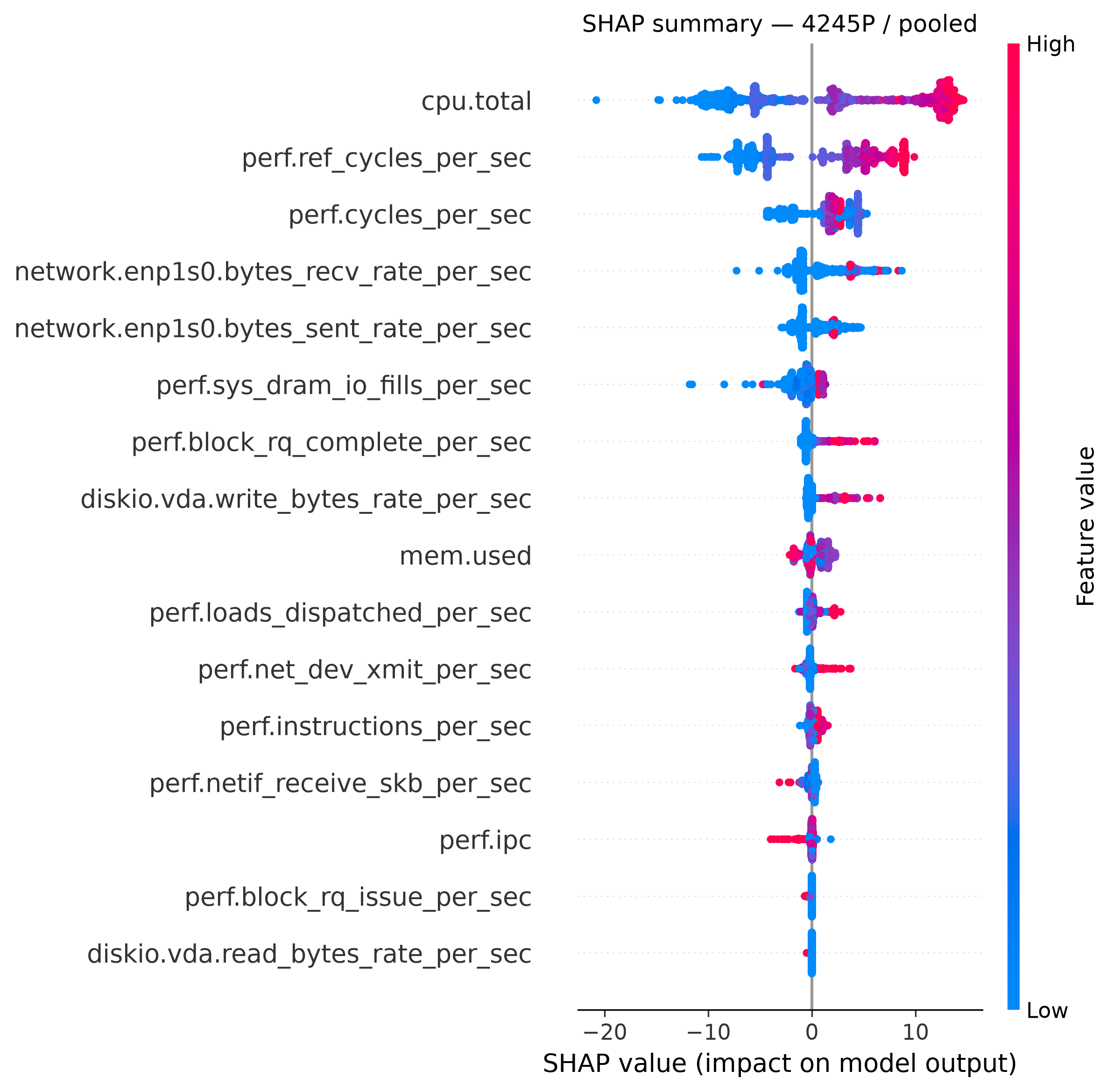}
    \label{fig:supp-shap-pooled-amd-epyc-4245p}
\end{figure}
 
\begin{figure}[H]
    \centering
    \caption{SHAP summary for the within-platform workload-pooled model on the AMD EPYC 7443P. Each point represents one held-out observation; horizontal position gives the SHAP contribution to predicted VM-associated power, and color denotes the corresponding feature value.}
    \includegraphics[width=0.82\textwidth]{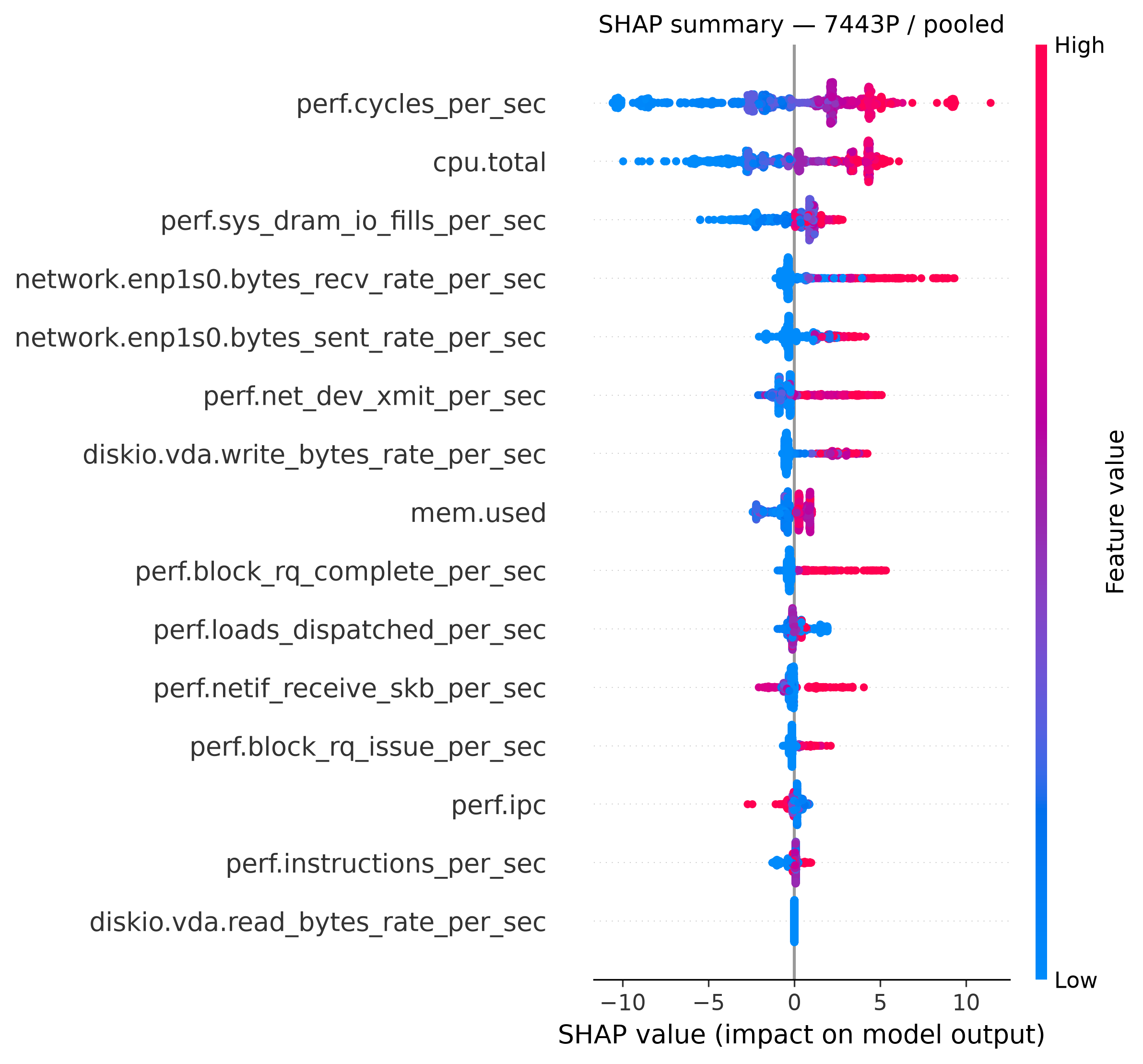}
    \label{fig:supp-shap-pooled-amd-epyc-7443p}
\end{figure}

\end{document}